\documentclass[final,5p,times,twocolumn]{elsarticle}

\usepackage{amssymb}
\usepackage{enumerate}
\usepackage{url}
\usepackage{multirow}
\usepackage{longtable}
\usepackage{caption}
\usepackage{amsmath}
\usepackage{makecell}
\usepackage{enumitem}

\usepackage[ruled,vlined,linesnumbered,noend]{algorithm2e}
\usepackage{colortbl}
\usepackage[american,USenglish,english]{babel}
\biboptions{numbers,sort&compress}
\newtheorem{definition}{Definition}
\newtheorem{example}{Example}

\journal{Knowledge-Based Systems}

\begin{document}

\begin{frontmatter}

\title{Ontology Revision based on Pre-trained Language Models}

\author{Qiu Ji$^{1,2}$}
\author{Xiaoping Zhang$^{3}$\corref{mycorrespondingauthor}\cortext[mycorrespondingauthor]{Corresponding author}}
\author{Yuxin Ye$^{4,2}$}
\author{Guilin Qi$^{5,6}$}
\author{Jiaye Li$^{7}$}
\author{Site Li$^{7}$}
\author{Jianjie Ren$^{8}$}
\author{Songtao Lu$^{8}$}

\address{
	{$^1$School of Modern Posts \& Institute of Modern Posts, Nanjing University of Posts and Telecommunications, Nanjing, P.R. China\\
	$^2$Key Laboratory of Symbolic Computation and Knowledge Engineering of Ministry of Education, Jilin University, Changchun, 130012, P.R. China\\
		qiuji@njupt.edu.cn\\

	$^3$National Data Center ofTraditional Chinese MedicineChina Academy of ChineseMedical Sciences,
		Beijing, 100700, P.R. China\\
		xiao\_ping\_zhang@139.com\\

	$^4$College of Computer Science and Technology, Jilin University, Changchun, 130012, P.R. China\\ 
		yeyx@jlu.edu.cn\\
		
	$^5$School of Computer Science and Engineering, Southeast University, Nanjing, China}\\
	$^6$Key Laboratory of Computer Network and Information Integration (Southeast University), Ministry of Education, Nanjing,  China\\
gqi@seu.edu.cn\\

	$^7$School of Mathematics, Southeast University, Nanjing, P.R. China\\
	lee559@seu.edu.cn, site-li@seu.edu.cn\\
	
	$^8$Chien-Shiung Wu College, Southeast University, Nanjing,  P.R.China\\
    seu\_rjj@seu.edu.cn, seu\_lst@seu.edu.cn\\
}

\begin{abstract}
	Ontology revision aims to seamlessly incorporate a new ontology into an existing ontology and plays a crucial role in tasks such as ontology evolution, ontology maintenance, and ontology alignment. Similar to repair single ontologies, resolving logical incoherence in the task of ontology revision is also important and meaningful, because incoherence is a main potential factor to cause inconsistency and reasoning with an inconsistent ontology will obtain meaningless answers.
	To deal with this problem, various ontology revision approaches have been proposed to define revision operators and design ranking strategies for axioms in an ontology. However, they rarely consider axiom semantics which provides important information to differentiate axioms. In addition, pre-trained models can be utilized to encode axiom semantics, and have been widely applied in many natural language processing tasks and ontology-related ones in recent years.
	Therefore, in this paper, we study how to apply pre-trained models to revise ontologies. We first define four scoring functions to rank axioms based on a pre-trained model by considering various information from an ontology. Based on the functions, an ontology revision algorithm is then proposed to deal with unsatisfiable concepts at once. To improve efficiency, an adapted revision algorithm is designed to deal with unsatisfiable concepts group by group. 
	We conduct experiments over 19 ontology pairs and compare our algorithms and scoring functions with existing ones. According to the experiments,  our algorithms could achieve promising performance. The adapted revision algorithm could improve the efficiency largely, and at most about 90\% of the time could be saved for some ontology pairs. Some of our scoring functions like \textsf{reliableOnt\_cos} could help a revision algorithm obtain better results in many cases, especially for those challenging ontology pairs like \textsf{OM8}.  We also provide discussion about the overall experimental results and guidelines for users to choose a suitable revision algorithm according to their semantics and performance.
\end{abstract}

\begin{keyword}
	Ontology revision \sep Inconsistency handling \sep Ontology matching \sep Pre-trained models\sep  Knowledge reasoning
\end{keyword}

\end{frontmatter}



\section{Introduction}\label{sec:intr}

Ontologies play a crucial role in the formal representation of knowledge. An ontology could define a set of entities including classes, properties or individuals, and it can also define axioms to describe the relationships among entities. 
After the Web Ontology Language (OWL) \footnote{\url{https://www.w3.org/TR/owl-overview/}} based on Description Logics (DLs) became a recommended specification of the W3C, especially with the development of knowledge graph \cite{Zhu23,Revello23}, ontologies play an increasingly important role. 
Currently, numerous OWL ontologies have been developed and applied in various fields like biological medicine \cite{NiuLPZ22}, public transportation \cite{RuckhausASC23} and financial area \cite{BunnellOY21}. Furthermore, ontologies
provide schema restrictions for knowledge graphs \cite{JiPCMY22} to facilitate their integration, querying, and maintenance. 
With the rigorous logical semantics provided by DLs, new logical consequences can be inferred from the axioms explicitly defined in an OWL ontology by applying a standard DL reasoner. Reasoning support is an essential characteristic of an OWL ontology. It requires that the ontology to be inferred is consistent. Otherwise, meaningless results will be obtained. A typical reason for inconsistency in DLs is due to unsatisfiable concepts which are interpreted as empty sets. An ontology is incoherent if it contains at least one unsatisfiable concept.
%
Incoherence often occurs when developing, maintaining and revising ontologies \cite{JohannaESWC10,LemboRSST17,JiSeu20,YuKbs21}. Resolving incoherence is critical to eliminate the potential inconsistency and make reasoning support work correctly.
To resolve unsatisfiable concepts in single ontologies, researchers have proposed various approaches  \cite{JiAs23,LiEswc23}. In this paper, we focus on resolving incoherence when revising ontologies. 

For the task of ontology revision, an original ontology should be revised consistently when a new ontology is received, namely adding a new ontology to an original one should not lead to any inconsistency or incoherence. It is assumed that both the original ontology and the new one need to be consistent and coherent. 
Ontology revision has broad application scenarios.
In the case of ontology alignment or ontology matching, two source ontologies are used to revise  an alignment between them \cite{li2023graph}, where an alignment consists of a set of mappings describing the relationships between entities from the two source ontologies. Each mapping can be translated to an OWL axiom \cite{JiLZQL22}.
Even for a single ontology, it can be divided into a static part and a rebuttal one, and then the static part can be used to revise the rebuttal one \cite{fu2016graph}.
So far, researchers have proposed various approaches to revising ontologies by deleting some axioms. One critical task is to decide which axioms should be removed for regaining the consistency or coherence of an ontology.
The work in \cite{qi2008kernel} proposed a kernel revision operator and chose axioms by considering their weights or frequencies.
The work in \cite{golbeck2009trust} also adopted the notions of kernel revision operators and incision functions to deal with inconsistency by utilizing trust information.
To improve efficiency, the authors in \cite{fu2016graph} proposed a graph-based method to revise DL-Lite ontologies, and ranked axioms according to their logical closures.
The work in \cite{JiSeu20} revised ontologies based on a partial order of axioms.
The method proposed in \cite{nikitina2012interactive} revised ontologies interactively and ranked axioms based on logical reasoning.
In the case of revising ontology mappings, the weights of mappings can also be exploited to differentiate the axioms in an alignment \cite{logmap,amlr,elog,pdlmv}.

Although existing ontology revision approaches provide various strategies or scoring functions to rank axioms, they rarely consider the semantics of axioms which provides an efficient way to differentiate axioms. 
To make use of the semantics of axioms, pre-trained models can be applied, because they can learn universal language representations on the large corpus and represent words in context \cite{QiuSun20}. 
Pre-trained models have been widely applied in many natural language processing tasks and ontology-related tasks in recent years and have achieved promising performance \cite{ptmSurvey2020,He22aaai,ma23bert}. Through a pre-trained model, ontology axioms can be encoded as vectors to preserve their semantics.
%
%
%

In this paper, we study how to revise ontologies based on pre-trained models. Specifically, ontology axioms need to be translated into natural language sentences first, and then a dense vector could be computed for each sentence based on a pre-trained model. With the obtained vectors, the similarity between any two axioms could be calculated by using a traditional distance metric like Cosine Distance. Based on the similarities, four scoring functions are defined for the ontology revision task. 
Afterwards, two concrete algorithms are designed to resolve the incoherence encountered when revising an ontology according to a new one. One algorithm needs to compute all minimal incoherence-preserving sub-ontologies (MIPS) first and then associates a score to each axiom in an obtained MIPS according to a scoring function. After that, a revision solution can be obtained based on the subsets extracted from MIPS according to axiom scores. The other algorithm deals with unsatisfiable concepts group by group to cope with the high computation cost of all MIPS.
To verify our algorithms and scoring functions, we compare our algorithms with existing algorithms over 19 pairs of ontologies. Each single ontology is coherent and consistent, but their combination is incoherent. Through the experiments, it is revealed that our ontology revision algorithms could achieve promising performances by utilizing the semantics of axioms. Finally, we discuss the overall experimental results and provide guidelines for users to choose a suitable revision algorithm.

The main contributions of this paper are summarized as follows:
\begin{itemize}
	\item Four scoring functions are designed for the task of ontology revision and rank axioms based on a pre-trained model to encode the semantics of axioms. These functions consider various aspects of an ontology to be revised and a new ontology to be combined. 	
	
	\item We propose two ontology revision algorithms. One needs to compute all MIPS for all unsatisfiable concepts, and the other deals with unsatisfiable concepts group by group. Both algorithms can be configured with various parameters such as  similarity measures and scoring functions.  	
	
	\item We implement and evaluate our algorithms over 19 ontology pairs by comparing them with existing ones. Experimental results reveal that our adapted algorithms could achieve much higher efficiency than the ones based on all MIPS, and at most about 90\% of the time has been saved. Additionally, our adapted algorithm \textsf{reliableOnt\_cos} has excellent performance in many cases, especially for those challenging ontologies. Finally, some guidelines are provided to users for choosing a specific revision algorithm.
\end{itemize}

The rest of the paper is organized as follows. Section \ref{sec:back} provides a brief introduction to background knowledge. Section \ref{sec:app} presents our approach to revising ontologies by providing some key definitions. Two concrete ontology revision algorithms are designed in Section \ref{sec:alg}. Section \ref{sec:exp} describes the experimental results. Related works are introduced in Section \ref{sec:related}, followed by conclusions and future works in Section \ref{sec:conclusion}.

\section{Background Knowledge}\label{sec:back}
This section {introduces} some preliminary of Description Logic ontologies and the basic definitions related to ontology revision. It also provides some basic knowledge about pre-trained models and similarity metrics.

\subsection{Description Logics ontologies}

A Description Logic (DL) ontology consists of a set of concepts, roles, or individuals. Individuals represent individual instances in a domain, concepts represent collections of instances in the domain, and roles represent binary relations between instances. Entities can be atomic concepts, atomic roles, or individuals. They can also be complex concepts or roles constructed by connecting entities using various constructors, such as existential quantifiers ($\exists$), universal quantifiers ($\forall$), conjunction ($\sqcap$), or disjunction ($\sqcup$).

A DL ontology can also define axioms to describe the relationships between entities. The axioms are typically divided into TBox axioms and ABox axioms. A TBox includes concept axioms and role axioms, which have the formats like $C \sqsubseteq D$ and $R \sqsubseteq S$, where $C$ and $D$ are concept descriptions, and $R$ and $S$ are role descriptions. An ABox  may include concept assertions and role assertions, which have formats like $C(a)$ and $R(a,b)$, where $a$ and $b$ are specific instances. 

\begin{example}\label{ex:onto}
Take an ontology including the following axioms as an example: 

$\phi_0:$ \textsf{MasterStudent $\sqsubseteq$ Student},

$\phi_1:$ \textsf{BachlorStudent $\sqsubseteq$ Student}, 

$\phi_2:$ \textsf{Judge $\sqsubseteq \neg$ Student}, 

$\phi_3:$ \textsf{Judge $\sqsubseteq$ Person},

$\phi_4:$ \textsf{StudentJudge $\sqsubseteq$ Student}, 

$\phi_5:$ \textsf{StudentJudge $\sqsubseteq$ Judge}, 

$\phi_6:$ \textsf{hasClassmate $\sqsubseteq$ hasRelation }, 

$\phi_7:$ \textsf{StudentJudge(Jackie)},

$\phi_8:$ \textsf{Student(Krissy)},

$\phi_9:$ \textsf{hasClassmate(Jackie, Krissy)}.\\
Among these axioms, the axiom $\phi_2$ represents that two concepts $Judge$ and $Student$ are disjoint. $\phi_6$ indicates that the role \textsf{hasClassmate} is a sub-role of \textsf{hasRelation}. $\phi_7$ means that the concept \textsf{StudentJudge} has an individual \textsf{Jackie}. Similarly, $\phi_8$ describes that the individual \textsf{Krissy} belongs to the concept \textsf{Student}. $\phi_9$ tells that the relation between two individuals \textsf{Jackie} and \textsf{Krissy} is \textsf{hasClassmate}.
Each of other axioms represent the subsumption relation between two atomic concepts.
For this ontology, its TBox includes the axioms from $\phi_0$ to $\phi_6$, and its ABox contains the axioms left.
\end{example}

Different DL languages may contain diverse constructors and axiom types. A basic DL language is $\mathcal{AL}$ which allows negation, union of atomic concepts, universal quantifiers and limited existential quantifiers. A more expressive DL language can be obtained by adding various factors to $\mathcal{AL}$, such as $\mathcal{N}$: quantity restriction, $\mathcal{I}$: inverse roles, $\mathcal{O}$: nominals, $\mathcal{D}$: datatypes and $\mathcal{H}$: role hierarchy. In addition, $\mathcal{S}$ is obtained by adding transitive roles to $\mathcal{ALC}$.

In DLs, concepts and roles correspond to classes and properties in OWL, respectively. DL-Lite is an important sub-language of OWL, and is specifically tailored to capture basic ontology languages while keeping all reasoning tasks tractable \cite{dl-lite}. Since the complexity of reasoning with DL-Lite ontologies is polynomial time, many inconsistency handling approaches were designed for DL-Lite ontologies.

\subsection{Ontology revision in DLs}

Before introducing the definitions related to ontology revision, we first provide those related to a single ontology.

\begin{definition}(\textbf{Inconsistent Ontology})\label{def:inconsistento}
	An ontology $K$ is inconsistent if and only if the model set of $K$ is empty.
\end{definition}
For inconsistent ontologies, the conclusions derived by standard DL reasoners are likely to be completely meaningless.
\begin{definition}(\textbf{Unsatisfiable Concept})\label{def:unsatisc}\cite{schlobach2003non}
	A named concept $C$ in an ontology $K$ is unsatisfiable if and only if for every model $I$ of $K$, $C^{\mathcal(I)}=\emptyset$. Otherwise, $C$ is satisfiable.
\end{definition}
Definition \ref{def:unsatisc} states that $C$ in $K$ is unsatisfiable if and only if it is interpreted as an empty set in every model of $K$. 
\begin{definition}(\textbf{Incoherent Ontology})\label{def:incoherento}\cite{schlobach2003non}
	An ontology $K$ is incoherent if and only if there exists at least one unsatisfiable named concept in $K$ .
\end{definition}
Since declaring an instance of an unsatisfiable concept would lead to inconsistency, incoherence is a potential factor to cause inconsistency. Usually, incoherence occurs in a TBox, while inconsistency is discussed across an entire ontology. In this paper, we only focus on incoherence and follow the definitions given in \cite{schlobach2003non}.

To explain why a concept is unsatisfiable, several or all  minimal unsatisfiability-preserving subsets (MUPS) can be computed.
\begin{definition} (\textbf{MUPS})\label{def:mups}
	Let $C$ be an unsatisfiable concept in an ontology $K$. A subset $K'\subseteq K$ is considered as a MUPS of $C$ in $K$ if $C$ is unsatisfiable in $K'$, but satisfiable in any $K''\subset K'$.
\end{definition}
A MUPS of $C$ is actually a minimal sub-ontology of $K$ to explain the unsatisfiability of $C$. In real-life incoherent ontologies, an unsatisfiable concept often has more than one MUPS.

To explain the incoherence of an ontology, a set of minimal incoherence-preserving subsets (MIPS) can be computed.
\begin{definition}(\textbf{MIPS})\label{def:mips}
	For an incoherent ontology $K$, its sub-ontology $K'{\subseteq}K$ is
	a MIPS of $K$ if $K'$ is incoherent and every sub-ontology $K''{\subset}K'$ is coherent.
\end{definition}
Obviously, a MIPS is a MUPS, and a MUPS includes all axioms of a MIPS. MUPS or MIPS can be used to resolve the unsatisfiability of a concept or the incoherence of an ontology.

\begin{example}
	For the ontology given in Example \ref{ex:onto}, it is incoherent and inconsistent. There is one unsatisfiable concept \textsf{StudentJudge} which has one MUPS $\{\phi_2,\ \phi_4,\ \phi_5\}$. This MUPS means that the concept \textsf{StudentJudge} has two disjoint super-concepts. Since there is only one MUPS, it is also a MIPS.
\end{example}

{
In the case of ontology revision, it is usually assumed that we have an ontology $K$ to be revised and a new ontology $K_0$ to be combined without making any changes. Each of the two ontologies is coherent and consistent while their combination is inconsistent or incoherent. In this paper, we focus on resolving incoherence since incoherence is a potential factor to cause inconsistency. Since $K$ is changeable and $K_0$ is unchangeable, we call $K$ as \textit{a rebuttal ontology} and $K_0$ as \textit{a reliable ontology}.
To resolve incoherence, existing works designed algorithms to delete or modify some axioms in $K$. We focus on deleting axioms. The task of ontology revision in this work can be formally defined as follows.}
\begin{definition}(\textbf{Ontology Revision})\label{def:ontRevision}
	Let $K$ be a rebuttal ontology and $K_0$ be a reliable ontology. Ontology revision is to remove a set of axioms $S$ from $K$ such that removing the axioms in $S$ from $K$ makes the union of $K_0$ and the modified $K$ coherent. Namely,
	 	
	$(K\setminus S)\cup K_0 \not\models C\sqsubseteq \bot\ $ for any $C\in K$.
\end{definition}
{
From Definition \ref{def:ontRevision} we can see that the selection of axioms in $S$ is the key to revise an ontology. Among existing ontology revision approaches, computing a set of axioms to be removed based on MIPS is a popular way \cite{qi2008kernel,fu2016graph,JiSeu20}. In this work, we also follow the idea. Since the computation of MIPS is often depend on all MUPS of all unsatisfiable concepts, we provide the formal definitions about MUPS and MIPS for ontology revision. For clarity, we use R-MUPS and R-MIPS to separately indicate MUPS and MIPS used in the task of ontology revision. }
%
\begin{definition}(\textbf{R-MUPS})\label{def:mupsk}
	Assume we have a reliable ontology $K_0$ and  an ontology $K$ to be revised. For an unsatisfiable concept $C$ in $K$ w.r.t. $K_0$, a R-MUPS $K'\subseteq K$ of $C$ w.r.t. $K_0$ satisfies: 
	
	(1) $C$ is unsatifiable in $K' \cup K_0$; 
	
	(2) For each $K''\subset K'$, $C$ is satisfiable in $K'' \cup K_0$. \\
	For convenience, the set of all R-MUPS of $C$ in $K$ w.r.t. $K_0$ is denoted by $MUPS_{K_0}(K, C)$.	
\end{definition}

\begin{definition}(\textbf{R-MIPS})\label{def:mipsk}
	Assume we have a reliable ontology $K_0$ and  an ontology $K$ to be revised. A R-MIPS $K'$ of $K$ w.r.t. $K_0$ is a subset of $K$ and satisfies the following conditions: 
	
	(1) $K' \cup K_0$ is incoherent; 
	
	(2) For each $K''\subset K'$, $K'' \cup K_0$ is coherent. \\
	The set of all R-MIPS of $K$ w.r.t. $K_0$ is denoted by $MIPS_{K_0}(K)$.	
\end{definition}

According to the definitions of R-MUPS and R-MIPS, we can see that they separately become MUPS and MIPS  when $K_0$ is an empty set. 

\begin{example}\label{ex:twoOnts}
	Following Example \ref{ex:onto}, assume we have a reliable ontology $K_0=\{\phi_0,\ \phi_1,\ \phi_2\}$ and a rebuttal ontology $K=\{\phi_3,\ \phi_4,\ \phi_5,\ \phi_6\}$. The concept \textsf{StudentJudge} in $K$ is unsatisfiable. It has a R-MUPS $M=\{\phi_4,\ \phi_5\}$, because it is unsatisfiable in $M\cup K_0$ but satisfiable in any $M'\cup K_0$ where $M'\subset M$. The R-MUPS $M$ is also a R-MIPS.
\end{example}

\subsection{Pre-trained models and similarity metrics}

Pre-trained models primarily leverage a large amount of unlabeled data available on the web for training, avoiding the high cost of supervised annotation. They could represent words or sentences with high-dimensional vectors in a semantic way, and thus have been widely accepted in both academic and industrial fields \cite{ptmSurvey2020,HousseinMA21,Bhargava022}. They can be used to perform various tasks without a resource-consuming training process. 
One of the most popular pre-trained models is BERT (Bidirectional Encoder Representations from Transformers) \cite{devlin2018bert}, which was proposed by Google and consists of a bidirectional encoder based on the Transformer architecture. 
BERT has achieved excellent results in various ontology or knowledge graph-related tasks \cite{He22aaai,ma23bert}. In this paper, BERT is adopted to obtain vectors of sentences due to its semantic representation.  

Before applying a pre-trained model, axioms need to be transformed into natural language sentences first. Similar to our previous work in \cite{JiAs23}, we use the tool of NaturalOWL\footnote{\url{http://www.aueb.gr/users/ion/publications.html}} which is described in the work of  \cite{AndroutsopoulosLG13}. After that, the sentences can be converted into vectors by applying a pre-trained model. With these vectors, the similarity between two sentences could be computed by exploiting a distance metric or similarity metric like Cosine Distance and Euclidean Distance. To be specific, the following definitions show how to compute a similarity value for two vectors based on a distance metric.

\begin{definition}(\textbf{sim}$_{cos}$)\label{def:sim-cos}
	The similarity metric based on Cosine Distance (marked as ${sim}_{cos}$) is formally defined as follows:
	\begin{equation*}
		\begin{aligned}
			sim_{cos}(v_1,v_2) & =\dfrac{1}{2}(1+\dfrac{v_1\cdot v_2}{||v_1||\times ||v_2||}) \\		
			& =\dfrac{1}{2}(1+\dfrac{\sum_{i=1}^d v_{1i}\times v_{2i}}{\sqrt{\sum_{i=1}^d (v_{1i})^2}\times \sqrt{\sum_{i=1}^d (v_{2i})^2}})
		\end{aligned}
	\end{equation*}
	Here, $v_{1i}$ and $v_{2i}$ indicate the $i$th element in the vectors $v_1$ and $v_2$ respectively. $||v_1||$ and $||v_1||$ indicate the norms of $v_1$ and $v_2$ separately. $d$ is the dimension of $v_{1}$ or $v_{2}$, both of them have the same dimension. 
\end{definition}

\begin{definition}(\textbf{sim}$_{euc}$)\label{def:sim_euc}
	The similarity metric based on Euclidean Distance (marked as ${sim}_{euc}$) is defined as follows:
	
	$sim_{euc}(v_1,v_2)=\dfrac{k}{k+\sqrt{\sum_{i=1}^d (v_{1i}-v_{2i})^2}}$\\
	Here, $k$ is a positive integer and $d$ indicates the dimension of $v_{1}$ or $v_{2}$. Both vectors have the same dimension. 
\end{definition}
According to this definition, we can observe that the greater $k$ is, the greater the similarities based on Euclidean Distance could reach.
Both similarity functions range from $0$ to $1$ since they have been normalized already. 

\section{Approach}\label{sec:app} 
In this section, we present our approach to resolving the incoherence  in the case of ontology revision. 

To resolve the incoherence in a rebuttal ontology $K$ w.r.t. a reliable ontology $K_0$ by removing some axioms from $K$, a natural way is to remove at least one axiom from each R-MIPS in ${MIPS}_{K_0}(K)$. In the following, we first define an incision function to choose axioms from R-MIPS.

\begin{definition}(\textbf{Incision Function})\label{def:incisionFunc}
	Assume we have a reliable ontology $K_0$ and a rebuttal ontology $K$. An incision function $\sigma$ for $K$ w.r.t. $K_0$ is a function from $2^{2^K}$ to $2^K$ and is defined as follows:
	
	(i) $\sigma(MIPS_{K_0}(K))\subseteq \bigcup_{M\in {MIPS}_{K_0}(K)} M$;
	
	(ii) if $M \in {MIPS}_{K_0}(K)$, then $M\cap \sigma(MIPS_{K_0}(K))\neq \emptyset$.
\end{definition}
In Definition \ref{def:incisionFunc}, the first condition indicates the axioms selected from $K$ by applying an incision function must belong to the union of all R-MIPS of $K$ w.r.t. $K_0$. The second one means the set of selected axioms must have an intersection with each R-MIPS. Overall, an incision function provides a general standard about how to choose axioms for removing to regain coherence.

As explained in our previous work in \cite{qi2008kernel}, an incision function is desired to be minimal for catering to the characteristic of minimal change. A minimal incision function can be formally defined as follows.

\begin{definition}(\textbf{Minimal Incision Function})\label{def:minIncisionFunc}
	Assume we have a reliable ontology $K_0$ and  a rebuttal ontology $K$. An incision function $\sigma$ for $K$ w.r.t. $K_0$ is minimal if there is no other incision function $\sigma'$ for $K$ such that $\sigma'(MIPS_{K_0}(K))\subset \sigma(MIPS_{K_0}(K))$.
\end{definition}
A minimal incision function selects a minimal set of axioms from each R-MIPS. Among all of the minimal incision functions, some of them may select more axioms than others. For example, if two minimal incision functions select axioms $\{a_1,\ a_2,\ a_3\}$ and $\{a_1,\ a_4\}$ separately, the first incision function chooses one more axiom than the second one. In the following, an incision function is defined to choose axioms with the minimal cardinality. 

\begin{definition}(\textbf{Cardinality-Minimal Incision Function})\label{def:carMinIncisionFunc}
	Assume we have a reliable ontology $K_0$ and  a rebuttal ontology $K$. An incision function $\sigma$ for $K$ w.r.t. $K_0$ is cardinality-minimal if there is no other incision function $\sigma'$ for $K$ such that $|\sigma'(MIPS_{K_0}(K))|< |\sigma(MIPS_{K_0}(K))|$.
\end{definition}
In Definition \ref{def:carMinIncisionFunc}, for a set $S$, $|S|$ indicates the number of elements in the set $S$, namely the cardinality of $S$. The definition shows that, for the given ontologies $K$ and $K_0$, the number of axioms chosen by a cardinality-minimal incision function is minimal.

Note that the resulting set of an incision function corresponds to a diagnosis, and that of a minimal (or cardinality-minimal) incision function corresponds to a minimal (or cardinality-minimal) diagnosis \cite{JiAs23}. Removing all axioms in a diagnosis from an ontology will regain coherence of the ontology.

With an incision function, a kernel revision operator can be formally defined as follows.
\begin{definition}(\textbf{Kernel Revision Operator})\label{def:operator}
	Assume we have a reliable ontology $K_0$,  a rebuttal ontology $K$, and an incision function $\sigma$. The kernel revision operator $\circ_\sigma$ for $K$ w.r.t. $K_0$ is defined below:	\\
	
	$K\circ_\sigma K_0=(K\setminus \sigma(MIPS_{K_0}(K))) \cup {K_0}.$
\end{definition}
Definition \ref{def:operator} means that a unique ontology can be obtained by removing those axioms selected by an incision function. According to the definition of an incision function, we know that the resulting ontology of such an operator must be coherent.

As we can see, it is a critical step to decide which axioms in a R-MIPS should be removed. Usually, the principle of a minimal change is desired when deleting information. It would be better to remove fewer axioms or remove some axioms with less information loss (e.g., weights or trusts) \cite{qi2008kernel,golbeck2009trust,li2023graph}. 
%
In this work, we consider the semantic information of axioms and provide the following scoring functions to rank each axiom in a R-MIPS. 

{When ranking an axiom, we consider the semantic relationship between it and any other axiom in a given set. The similarity between two axioms can be measured by the similarity metrics ${sim}_{cos}$ or ${sim}_{euc}$ based on a pre-trained model (see Definition \ref{def:sim-cos} and \ref{def:sim_euc}). The low similarity between two axioms usually means that they have weak semantic relationship, or they have no semantic relationship at all. To reduce the influence of low similarity values, a threshold is used and only those similarity values over the threshold are considered. }
\begin{definition}(\textbf{Similarity between an axiom set and an axiom})\label{def:sim}
	Assume we have a reliable ontology $K_0$ and a rebuttal ontology $K$. Given an axiom $\alpha$ in $K$, an axiom set $S\subseteq K\cup K_0$ and a predefined threshold $t$, we define the similarity between $S$ and  $\alpha$ with respect to $t$ as follows:\\
	
	$sim_{K_0}^K(S, \alpha, t) = \dfrac{1}{|S'|+1}\sum_{\beta \in S'}sim(v_{\alpha},v_{\beta}), $\\ \\
	where $S'=\{\beta\in S | sim(v_{\alpha},v_{\beta})>=t\}$.
\end{definition}
{
Definition \ref{def:sim} computes the average similarity between an axiom $\alpha$ and all axioms in an axiom set $S$ with a threshold $t$. Firstly, it extracts a subset $S'$ from $S$, so that the similarity between each axiom in $S'$ and $\alpha$ is not less than $t$. Then, the average similarity between $\alpha$ and an axiom in S' is calculated.}
This definition is similar to the threshold-based degree given in our previous work \cite{JiAs23}. Their main difference is that we add $1$ to $|S'|$ for smoothing our scoring function. In this way, we could avoid the case that denominator equals zero  when $S'$ is an empty set. 

Based on Definition \ref{def:sim}, a scoring function could be defined based on all R-MIPS. Since the axioms in a R-MIPS are often regarded as problematic information, an axiom in a R-MIPS would be more problematic if it had higher similarity with other axioms in R-MIPS. Definition \ref{def:score_mipsUnion} realizes this idea.
\begin{definition}(\textbf{Scoring function based on MIPS union})\label{def:score_mipsUnion}
	Assume we have a reliable ontology $K_0$ and a rebuttal ontology $K$. For an axiom $\alpha$ in $K$ and a predefined threshold $t$, the score of $\alpha$ based on R-MIPS union can be defined:\\
	
	$score_{mipsUnion}(K,K_0,\alpha, t) = sim_{K_0}^K(\bigcup_{M \in MIPS_{K_0}(K)}M, \alpha, t) $.
\end{definition}
This definition roughly calculates the average similarity between the axiom $\alpha$ and an axiom in the union of all R-MIPS. 

Alternatively, for an axiom to be ranked, we could consider the relationship between it and each R-MIPS. 
\begin{definition}(\textbf{Scoring function based on MIPS})\label{def:score_mips}
	Assume we have a reliable ontology $K_0$ and a rebuttal ontology $K$. For an axiom $\alpha$ in $K$ and a predefined threshold $t$, the score of $\alpha$ based on R-MIPS can be defined as follows:\\	
	$score_{mips}(K,K_0,\alpha,t) = \dfrac{1}{|MIPS_{K_0}(K)|}\sum_{ M \in MIPS_{K_0}(K) } sim_{K_0}^K(M,\alpha,t) $.
\end{definition}
The definition first calculates the similarity between the axiom to be ranked and a R-MIPS, and then average all of the obtained similarities.  

Similarly, since all axioms in a rebuttal ontology are unreliable, an axiom should have a high priority to be removed if it has a high similarity with the rebuttal ontology. Thus, we obtain another scoring function $score_{rebuttalOnt}$.
\begin{definition}(\textbf{Scoring function based on a rebuttal ontology})\label{def:score_rebuttalOnt}
	Assume we have a reliable ontology $K_0$ and a rebuttal ontology $K$. For an axiom $\alpha$ in $K$ and a predefined threshold $t$, the score of $\alpha$ based on $K$ can be defined as follows:\\
	
	$score_{rebuttalOnt}(K,K_0,\alpha, t) = sim_{K_0}^K(K, \alpha, t) $.
\end{definition}
The definition ranks an axiom with the average similarity between the axiom and an axiom in a rebuttal ontology.

Additionally, because all axioms in a reliable ontology are stable and reliable and it is often desired to have a compact ontology containing axioms that are semantically connected, an axiom should have a high priority to be removed if it has a low similarity with the reliable ontology. To realize this idea, we obtain the following scoring function.
%
%
\begin{definition}(\textbf{Scoring function based on a reliable ontology})\label{def:score_reliableOnt}
	Assume we have a reliable ontology $K_0$ and an ontology $K$ to be revised. For an axiom $\alpha$ in $K$ and a predefined threshold $t$, we define its score:\\
	
	$score_{reliableOnt}(K,K_0,\alpha, t) = sim_{K_0}^K(K_0,\alpha,t) $.
\end{definition}
The definition ranks an axiom with the average similarity between an axiom and an axiom in the reliable ontology $K_0$.

\begin{example}\label{ex:computeScores}
	Following the two ontologies given in Example \ref{ex:twoOnts}, we provide the process of ranking axioms in a R-MIPS by taking the scoring function $score_{reliableOnt}$  with the similarity metric $sim_{cos}$  as an example. 
	
	For R-MIPS $\{\phi_4,\ \phi_5\}$ in $K$ w.r.t. $K_0$, we rank the axiom $\phi_4$ first. After applying the pre-trained model BERT to compute vectors for the axioms in $K$ and $K_0$, we obtain: $sim_{cos}(\phi_4,\ \phi_0)\approx 0.81$, $sim_{cos}(\phi_4,\ \phi_1)\approx 0.78$ and $sim_{cos}(\phi_4,\ \phi_2)\approx 0.74$. Assuming $t=0.5$, we obtain the score of $\phi_4$:
	
	$score_{reliableOnt}(K,K_0,\phi_4, 0.5)=sim_{K_0}^K(K_0,\phi_4,0.5)$
		
	$\ \ \ \ \ \ =\frac{1}{3+1}(sim_{cos}(\phi_4,\ \phi_0)+sim_{cos}(\phi_4,\ \phi_1)+sim_{cos}(\phi_4,\ \phi_2))$
	
	$\ \ \ \ \ \ \approx \frac{1}{4}(0.81+0.78+0.74) \approx 0.58$.\\
	Similarly, we have  $sim_{cos}(\phi_5,\ \phi_0)\approx 0.61$, $sim_{cos}(\phi_5,\ \phi_1)\approx 0.54$ and $sim_{cos}(\phi_5,\ \phi_2)\approx 0.56$. The score of $\phi_5$ can be calculated:
	
	$score_{reliableOnt}(K,K_0,\phi_5, 0.5)$
	
	$\ \ \ \ \ \ =\frac{1}{3+1}(sim_{cos}(\phi_5,\ \phi_0)+sim_{cos}(\phi_5,\ \phi_1)+sim_{cos}(\phi_5,\ \phi_2))$
		
	$\ \ \ \ \ \ \approx \frac{1}{4}(0.61+0.54+0.56) \approx 0.43$.\\	
\end{example}
%
%
\section{Algorithm}\label{sec:alg}

In this section, we design two specific algorithms for ontology revision based on the scoring functions defined in the previous section. Before introducing the algorithms, we first present the details to compute a diagnosis for a  set of R-MIPS.

\begin{algorithm}[t]
	\KwData{A rebuttal ontology $K$, a reliable ontology $K_0$ and a set of R-MIPS $\mathcal{M}$ of $K$ w.r.t. $K_0$}
	\KwResult{A diagnosis $D$ to resolve $\mathcal{M}$}
	\Begin{
		$\mathcal{C}=\emptyset$   \\
		// Assign a score to each axiom in a R-MIPS\\
		$s(\gamma)=score_{mips}(K,K_0,\gamma)$ for $\gamma\in\bigcup_{M\in\mathcal{M}}M$\\
		// Extract subsets from R-MIPS\\
		\For{$M \in \mathcal{M}$}{
			$A = \{\alpha\in M : \not\exists \beta \in M, s(\beta)>s(\alpha)\}$  \\
			$\mathcal{C} = \mathcal{C} \cup \{ A \}  $ \\
		}
		$D' = \textsf{ILP}(\mathcal{C})$  \\	
		\Return{$D$}
	}
	\caption{Compute a diagnosis for a set of R-MIPS (i.e., \textsf{ComputeDiagnosis}($K,K_0,\mathcal{M}$)) }\label{alg:computeDiagnosis}
\end{algorithm}

Algorithm \ref{alg:computeDiagnosis} describes the steps to compute a diagnosis. It takes a reliable ontology $K_0$, a rebuttal ontology $K$ and a set of R-MIPS as inputs, and outputs a diagnosis to resolve all inputted R-MIPS. When computing R-MIPS, an existing incoherence-detecting algorithm (see \cite{JiLZQL22}) could be used.
In the algorithm, a scoring function should be first selected to assign scores to the axioms in the union of all R-MIPS. In this algorithm, the scoring function $score_{mips}$ is selected (see Line 4) . 
From Line 6 to Line 8, the algorithm chooses a subset from each R-MIPS by considering those axioms with the highest score.
Line 9 computes a diagnosis over the extracted subsets by applying an integer linear programming (abbreviated as ILP) solver which is similar to our previous algorithm given in \cite{Ji19access}. The ILP-based method \textsf{ILP}($\mathcal{C}$) first associates a binary variable to each axiom in the union of the elements in $\mathcal{C}$, and then constructs an objective function over these variables. For each element in $\mathcal{C}$, a constraint is constructed. An optimal assignment is finally generated such that all constraints are satisfied, and it can be translated to DL axioms easily.

It is noted that, the algorithm
invokes scoring function  $score_{mips}$  to rank axioms (see Line 3), which can be replaced by scoring functions $score_{mipsUnion}$ or $score_{rebuttalOnt}$.
If $score_{reliableOnt}$ is applied, Line 7 should be modified by changing greater-than symbol to less-than symbol. Namely, we choose those axioms with the lowest score from each R-MIPS. 
In the following, an example is provided to illustrate the step of subset extraction.
\begin{example}
	Following Example \ref{ex:computeScores}, we illustrate how to extract a subset from a R-MIPS when $score_{reliableOnt}$ is applied in Algorithm \ref{alg:reviseAlg}. For the R-MIPS $\{\phi_4,\ \phi_5\}$, the subset $\{\phi_5\}$ is extracted from it since $score_{reliableOnt}(K,K_0,\phi_5,0.5) < score_{reliableOnt}(K,K_0,\phi_4,0.5)$.
\end{example}

\begin{algorithm}[t]
	\KwData{A rebuttal ontology $K$ and a reliable ontology $K_0$}
	\KwResult{A diagnosis $D$ for $K$}
	\Begin{
		$\mathcal{M}=MIPS_{K_0}(K)$\\
		$D' = $\textsf{ComputeDiagnosis}($K,K_0,\mathcal{M}$)  \\
		// Minimize a diagnosis\\
		$D = D'$\\
		\For{$\alpha \in D' $}{
			\If{(($K\cup K_0)\setminus D) \cup \{\alpha\}$ is coherent}{
				$D = D \setminus \{\alpha\}$\\
			} 
		}
		\Return{$D$}
	}
	\caption{An ontology revision algorithm based on all R-MIPS}\label{alg:reviseAlg}
\end{algorithm}

Based on Algorithm \ref{alg:computeDiagnosis}, we propose an algorithm to revise an ontology based on all R-MIPS (see Algorithm \ref{alg:reviseAlg}). It takes a reliable ontology $K_0$ and a rebuttal ontology $K$ as inputs, and outputs a diagnosis to resolve the incoherence in $K$ w.r.t. $K_0$.
In this algorithm, all R-MIPS are computed first and a diagnosis can be calculated by invoking Algorithm \ref{alg:computeDiagnosis} (see Lines 2-3). Although the obtained diagnosis is minimal with respect to the subsets obtained in Algorithm \ref{alg:computeDiagnosis}, it may not be minimal regarding the R-MIPS. 
For example, assume we have two R-MIPS $\{a_1:\ 0.6,\ a_2:\ 0.5\}$ and $\{a_1:\ 0.6,\ a_3:\ 0.7\}$, and the corresponding subsets extracted by an extraction strategy are $\{a_2:\ 0.5\}$ and $\{a_1:\ 0.6\}$ separately, where $a_i$ (i=1, 2, 3) indicates an axiom and a real number like 0.5 or 0.6 represents a weight. The final diagnosis is $\{a_1,\ a_2\}$. In the set, $a_2$ is actually redundant since removing $a_1$ has resolved the two R-MIPS already.
%
Therefore, the algorithm checks all axioms in the diagnosis $D'$ and removes those redundant axioms (see Lines 5-8 in Algorithm \ref{alg:reviseAlg}). Finally, a minimal diagnosis $D$ is obtained.
By removing all axioms in $D$ from $K$, the union of modified $K$ and $K_0$ becomes coherent.

\begin{algorithm}[t]
	\KwData{A rebuttal ontology $K$, a reliable ontology $K_0$ and a step length $n$}
	\KwResult{A diagnosis $G$ for $K$}
	\Begin{
		$\mathcal{MU}, D=\emptyset$   \\
		$ k = 0$\\
		$UC$ = All unsatisfiable concepts in $K$ w.r.t. $K_0$  \\	
		\For{$C \in UC$}{
			\If{$C$ is satisfiable in $K$ w.r.t. $K_0$}{continue}			
			$\mathcal{M}_C$ = $MUPS_{K_0}(K,C)$  \\
			$\mathcal{MU} = \mathcal{MU} \cup \mathcal{M}_C  $ \\
			$k=k+1$\\
			\If{$k==n$}{	
				$\mathcal{M} = $ Local R-MIPS based on $\mathcal{MU}$\\		
				$D' =  \textsf{ComputeDiagnosis}(K, K_0, \mathcal{M})$  \\
				$D=D \cup D'$\\
				$K = K\setminus D'$\\
				$\mathcal{MU}=\emptyset$   \\
				$k=0$\\
			}
		}	
		\If{$\mathcal{MU}\neq\emptyset$}{
			$\mathcal{M} = $ Local R-MIPS based on $\mathcal{MU}$\\		
			$D' =  \textsf{ComputeDiagnosis}(K, K_0, \mathcal{M})$  \\
			$D=D \cup D'$\\
		}	
		// Minimize a diagnosis\\
    	$G = D$\\
    	\For{$\alpha \in D $}{
	    	\If{(($K\cup K_0)\setminus G) \cup \{\alpha\}$ is coherent}{
		    	$G = G \setminus \{\alpha\}$\\
		} 
	}
		\Return{$G$}
	}
	\caption{An adapted ontology revision algorithm}\label{alg:adaptedReviseAlg}
\end{algorithm}

Since computing all R-MIPS is often time-consuming or memory-consuming \cite{JiLZQL22}, it may be infeasible to obtain all of them within limited resources. To deal with this problem, our previous work in \cite{qi2008kernel} proposed an adapted revision algorithm to resolve unsatisfiable concepts one by one. 
However, it may not be necessary to revise ontologies in this way since computing all R-MUPS for an unsatisfiable concept is much easier due to their small sizes. 
Therefore, we design a trade-off revision algorithm to deal with unsatisifable concepts group by group. Namely, a set of local R-MIPS can be obtained over all R-MUPS of all unsatisfiable concepts in a group, and then we focus on ranking the axioms in local R-MIPS. In this way, a local revision solution may be obtained.
A local R-MIPS can be formally defined as follows:

%
\begin{definition}(\textbf{Local R-MIPS})\label{def:localIncMips}
	For a reliable ontology $K_0$ and a rebuttal ontology $K$, assume we have a set of unsatisfiable concepts $S$ in $K$ w.r.t. $K_0$. A sub-ontology $K'{\subseteq}K$ is
	a local R-MIPS of $K$ w.r.t. $K_0$ if the following conditions hold:
	(1) There exists a concept $C$ in $S$ such that $C$ is unsatisfiable in $K'$ w.r.t. $K_0$; (2) Each concept in $S$ is satisfiable in every sub-ontology $K''{\subset}K'$ w.r.t. $K_0$.
\end{definition}
Different with a global R-MIPS, a local R-MIPS is calculated based on a set of unsatisifalbe concepts in a rebuttal ontology with respect to a reliable one while not based on all unsatisifalbe concepts. Thus, a local R-MIPS is a R-MUPS, but it may not be a global one.

Based on the definition of local R-MIPS, we design an adapted algorithm to resolve all unsatisfiable concepts group by group (see Algorithm \ref{alg:adaptedReviseAlg}). The inputs of the algorithm include an ontology $K$ to be revised, a reliable ontology $K_0$ and a step length $n$ to deal with a fixed number of unsatisfiable concepts for each iteration. Its output is a diagnosis to resolve all unsatisfiable concepts in $K$ w.r.t. $K_0$.

\begin{table*}[t]
	\begin{center}
		\begin{footnotesize}
			\begin{tabular}{|l|l|r|r|r|r|r|r|r|r|}
				\hline
				\multicolumn{2}{|c|}{Ontology Names} &$|O|$&SubClass&EquClass&DisjClass&SubProp&Domain&Range\\
				Full names&Short names&&&&&&&\\
				\hline
				\multicolumn{9}{|c|}{Single ontologies coming from ontology alignment task}\\
				\hline
				ALOD2Vec-cmt-ekaw&M0&6&0&4&0&0&0&0\\
				ALOD2Vec-conference-ekaw&M1&26&0&24&0&0&0&0\\
				GMap-conference-edas&M2&22&0&13&0&0&0&0\\
				GMap-edas-ekaw&M3&12&0&10&0&0&0&0\\
				Lily-cmt-conference&M4&12&0&12&0&0&0&0\\
				Lily-edas-ekaw&M5&18&0&18&0&0&0&0\\
				OTMapOnto-cmt-conference&M6&38&0&27&0&0&0&0\\
				OTMapOnto-conference-edas&M7&35&0&27&0&0&0&0\\
				OTMapOnto-ekaw-sigkdd&M8&50&0&47&0&0&0&0\\
				TOM-iasted-sigkdd&M9&14&0&11&0&0&0&0\\
				\hline
				cmt-conference&O0&511&91&14&41&13&123&123\\
				cmt-ekaw&O1&459&118&1&101&8&83&83\\
				conference-edas&O2&1024&150&20&421&13&114&114\\
				conference-ekaw&O3&518&143&13&88&21&88&88\\
				edas-ekaw&O4&972&177&7&481&8&74&74\\
				ekaw-sigkdd&O5&349&134&7&74&8&51&51\\
				iasted-sigkdd&O6&474&296&23&1&0&68&68\\
				\hline
				\multicolumn{9}{|c|}{Single ontologies coming from km1500}\\
				\hline
				km1500-1000-1&km0&1000&729&0&176&0&39&56\\
				km1500-1000-2&km1&1000&730&0&184&0&48&38\\
				km1500-1000-3&km2&1000&731&0&163&0&53&53\\
				km1500-1000-4&km3&1000&733&0&170&0&52&45\\
				km1500-1000-5&km4&1000&736&0&174&0&41&49\\
				km1500-1000-6&km5&1000&749&0&168&0&43&40\\
				km1500-1000-7&km6&1000&723&0&190&0&50&37\\
				km1500-1000-8&km7&1000&737&0&176&0&46&41\\
				km1500-1000-9&km8&1000&774&0&149&0&36&41\\
				km1500-1000-10&km9&1000&722&0&168&0&53&57\\
				\hline
			\end{tabular}
			\vspace{3mm}
			\caption{Single ontologies used in our experiments, where $|O|$ indicates the number of axioms in an ontology $O$ and the last six columns present the number of axioms in a specific form. \label{tab:onts}}
		\end{footnotesize}
	\end{center}
\end{table*}

In Algorithm \ref{alg:adaptedReviseAlg}, all unsatisfiable concepts need to be calculated first  by using a standard DL reasoner such as Pellet \cite{SirinPGKK07}, and then iterates on all of these concepts (see lines 4-5). For each unsatisfiable concept, if it is still unsatisfiable in the modified ontology $K$ w.r.t. $K_0$, all of its R-MUPS will be computed (see lines 6-9). We use a variable $k$ to control the number of unsatisfiable concepts to be dealt with (see Line 10). If $k$ reaches the predefined length $n$, all local R-MIPS w.r.t. the $n$ unsatisfiable concepts will be computed based on all found R-MUPS (see lines 11-12). Once a set of local R-MIPS is obtained, a local diagnosis can be computed by invoking the algorithm \textsf{ComputeDiagnosis} (i.e., Algorithm \ref{alg:reviseAlg}), and the global diagnosis $D$ should be updated by adding all elements in $D'$ (see lines 13-14). Afterwards, $K$ needs to be updated by removing all axioms in $D'$, and the set of R-MUPS $\mathcal{MU}$ and the counter $k$ should be reset (see lines 15-17).
When all unsatisfiable concepts in $UC$ have been checked, the ``for" loop will be terminated.  Outside the loop, $\mathcal{MU}$  may not be empty as less than $n$ unsatisfiable concepts may have not been resolved. in such a case, all local R-MIPS are computed based on $\mathcal{MU}$, and a local diagnosis is calculated (see lines 18-20). The global diagnosis $D$ should be updated again (see Line 21).
Finally, after removing redundant axioms (see Lines 23-26), a  minimal diagnosis $G$ is obtained, and removing all axioms in $G$ from $K$ will regain coherence.

\section{Experiments}\label{sec:exp}

In this section, we first introduce the data set and experimental settings, and then provide experimental results.

It should be noted that all algorithms were implemented with OWL API\footnote{\url{http://owlcs.github.io/owlapi/}} in Java. 
The functionality of computing vectors for sentences  was implemented in Python,  and the pre-trained model BERT was applied to calculate vectors (see the introduction in Section \ref{sec:back}). To perform standard reasoning tasks, the widely used DL reasoner Pellet \cite{SirinPGKK07} was selected.
In addition, according to our experience, the parameter $k$ in the definition of {sim}$_{euc}$ (see Definition \ref{def:sim_euc}) was set to be $15$ for our experiments, and the threshold $t$ in the definition of similarity between an axiom set and an axiom (see Definition \ref{def:sim}) was assigned to be $0.5$.
All implementations together with our data set and experimental results are available online\footnote{{\url{https://github.com/QiuJi345/ontRevision}}}.

\subsection{Data set}
The data set consists of two groups of single ontologies (see Table \ref{tab:onts}). One group comes from the conference track on the platform of Ontology Alignment Evaluation Initiative (OAEI) \footnote{\url{http://oaei.ontologymatching.org/2021/conference/index.html}}, which started in 2004 and is to evaluate  ontology matching systems from various dimensions. In this group, a rebuttal ontology (see ontologies from M0 to M9) is an alignment between two single ontologies, and its full name is named with the names of a matching system and two single ontologies. A reliable ontology (see ontologies from O0 to O6) combines two single ontologies. 
For example, the reliable ontology \textsf{cmt-ekaw} combines single ontologies \textsf{cmt} and \textsf{ekaw}. The rebuttal ontology \textsf{ALOD2Vec-cmt-ekaw} indicates the alignment between \textsf{cmt} and \textsf{ekaw} which is generated by the matching system \textsf{ALOD2Vec} \cite{ALOD2Vec}. Similarly, other rebuttal ontologies are produced by ontology matching systems
\textsf{GMap} \cite{GMap},  Lily\cite{Lily},  OTMapOnto \cite{OTMapOnto} and TOM  \cite{TOM} separately. 
%
%
According to the matching results provided by OAEI 2021, 16 systems have participated the conference track, and the alignments produced by half of them contain unsatisfiable concepts. Among the 42 incoherent alignments, we chose 10 incoherent alignments and obtained 10 ontology pairs (see pairs from OM0 to OM9 in Table \ref{tab:ontPairs}) with different numbers of unsatisfiable concepts and various sizes of axioms. 

The other group comes from the consistent but incoherent ontology \textsf{km1500} that was learned on a text corpus consisting of the abstracts from the ``knowledge management" information space of the BT Digital Library \cite{text2onto}. The original ontology \textsf{km1500} contains more than 10,000 axioms and 1,000 unsatisfiable concepts. It is very challenging to resolve all unsatisfiable concepts. In our experiments, 10 coherent sub-ontologies were extracted (see ontologies from km0 to km8 in Table \ref{tab:onts}), and 9 ontology pairs were constructed for revision (see pairs from KM0 to KM9 in Table \ref{tab:ontPairs}). For each pair, the combination of its contained ontologies is incoherent.

Comparing the two groups of single ontologies or ontology pairs, it can be observed that each rebuttal ontology in the first group mainly contains axioms representing equivalent classes (i.e., EquClass in Table \ref{tab:onts}), and no more than 50 axioms are included in such an ontology. The corresponding ontology pairs usually contain less than 1,000 axioms in total, but the number of contained unsatisfiable concepts varies from 5 to 84.
Each ontology pair in the second group contains 2,000 axioms. Such ontologies only contain subsumptions (i.e., \textsf{SubClass} in Table \ref{tab:onts}), axioms representing equivalent classes, domain and range. Their expressivity is $\mathcal{ALC}$. They are much less expressive than all ontology pairs in the first group, and vary greatly in number of unsatisfiable concepts (ranging from 4 to 120).

\begin{table}[t]
	\begin{center}
		\begin{footnotesize}
			\begin{tabular}{|l|l|l|r|r|r|}
				\hline
				\multicolumn{2}{|c|}{Single Ontologies} & \multicolumn{3}{c|}{Ontology Pairs} & UCs\\
				Reliable ($K_0$) & Rebuttal ($K$) & Name & Expressivity & Size &   \\
				\hline\hline
				O1 & M0 & OM0  & $\mathcal{SHIN(D)}$  & 465  & 17  \\
				O3 & M1 & OM1  &  $\mathcal{SHIN(D)}$  & 544  & 34\\
				O2 & M2 & OM2  & $\mathcal{ALCHOIN(D)}$  & 1046  & 11  \\
				O4 & M3 & OM3  & $\mathcal{SHOIN(D)}$  & 984  & 23  \\
				
				O0 & M4 & OM4  & $\mathcal{ALCHIN(D)}$  & 523  & 6 \\
				O4 & M5 & OM5  & $\mathcal{SHOIN(D)}$  & 990  & 10  \\
				O0 & M6 & OM6  & $\mathcal{ALCHIN(D)}$  & 549  & 46   \\
				O2 & M7 & OM7  & $\mathcal{ALCHOIN(D)}$  & 1059  & 41  \\
				O5 & M8 & OM8  & $\mathcal{SHIN(D)}$  & 399  & 84\\				
				O6 & M9 & OM9  & $\mathcal{ALCHIN(D)}$ & 488  &   5  \\
				\hline\hline
				km0 & km1 & KM0 & $\mathcal{ALC}$ & 2000 &   4 \\
				km1 & km2 & KM1 & $\mathcal{ALC}$ & 2000 &   20 \\
				km2 & km3 & KM2 & $\mathcal{ALC}$ & 2000 &   33 \\
				km3 & km4 & KM3 & $\mathcal{ALC}$ & 2000 &   76 \\
				km4 & km5 & KM4 & $\mathcal{ALC}$ & 2000 &   29 \\
				km5 & km6 & KM5 & $\mathcal{ALC}$ & 2000 &   42 \\
				km6 & km7 & KM6 & $\mathcal{ALC}$ & 2000 &   120 \\
				km7 & km8 & KM7 & $\mathcal{ALC}$ & 2000  &   79 \\
				km8 & km9 & KM8 & $\mathcal{ALC}$ & 2000 &   20 \\
				
				\hline
			\end{tabular}
			\vspace{3mm}
			\caption{Information about ontology pairs in the task of ontology revision, where ``UCs" indicates unsatisfiable concepts. \label{tab:ontPairs}}
		\end{footnotesize}
	\end{center}
\end{table}

\subsection{Experimental Settings}\label{sec:setting}
All experiments were performed on a laptop with 1.99 GHz Intel$^{\textregistered}$ Core$^{TM}$ CPU and 16GB RAM, using a 64-bit Windows 11 operating system. A time limit of 1,000 seconds is set to compute R-MUPS for an unsatisfiable concept or compute a diagnosis. A black-box algorithm implemented in \cite{JiSeu20} was exploited to compute R-MUPS or R-MIPS.

We evaluate our two revision algorithms by using different subset extraction strategies configured in the following ways. 
\begin{itemize}
	\item \textsf{mipsUnion\_cos / mipsUnion\_euc}: The two extraction strategies indicate ranking axioms by the scoring function $score_{mipsUnion}$ (see Definition \ref{def:score_mipsUnion}) with similarity measures ${sim}_{cos}$ and ${sim}_{euc}$  (see Definition \ref{def:sim-cos} and Definition \ref{def:sim_euc}) separately, and then selecting axioms with the highest score from each R-MIPS.
	
	\item \textsf{mips\_cos / mips\_euc}: The two extraction strategies represent ranking axioms by the scoring function $score_{mips}$ (see Definition \ref{def:score_mips}) with similarity measures ${sim}_{cos}$ and ${sim}_{euc}$ separately, and then selecting axioms with the highest score from each R-MIPS.
	
	\item \textsf{rebuttalOnt\_cos / rebuttalOnt\_euc}: The two extraction strategies indicate ranking axioms by the scoring function $score_{rebuttalOnt}$ (see Definition \ref{def:score_rebuttalOnt}) with similarity measures ${sim}_{cos}$ and ${sim}_{euc}$ separately,  and then selecting axioms with the highest score from each R-MIPS.
	
	\item \textsf{reliableOnt\_cos / reliableOnt\_euc}: The two extraction strategies indicate ranking axioms by the scoring function $score_{reliableOnt}$ (see Definition \ref{def:score_reliableOnt}) with similarity measures ${sim}_{cos}$ and ${sim}_{euc}$ separately, and then selecting axioms with the lowest score from each R-MIPS.
\end{itemize}

Additionally, four existing extraction strategies below were chosen to compare with ours because they were frequently used in the existing works \cite{JiAs23}. 
\begin{itemize}
	\item \textsf{ex-base}: This is a baseline strategy to compute a diagnosis based on all R-MIPS directly without extracting subsets \cite{Ji19access}. Since the diagnosis found by this strategy is minimal regarding to all inputted R-MIPS, it is not necessary to check redundancy of axioms.
	
	\item \textsf{ex-score}: The strategy ranks axioms with their frequency \cite{qi2008kernel}. Namely, for an axiom, its frequency is the number of R-MIPS containing it. The strategy chooses the axioms with the highest frequency from each R-MIPS.
	
	\item \textsf{ex-shapley}: This strategy assigns a penalty to an axiom $\alpha$ in a R-MIPS, where the penalty is inversely proportional to the size of a R-MIPS where  $\alpha$ is contained \cite{TeymourlouieZNT18}. Thus, the strategy regards those axioms with the highest penalty score in a R-MIPS as candidates to remove.
	
	\item \textsf{ex-sig}: It is a signature-based strategy that originally ranks an axiom by summing the reference counts in other axioms for all entities appearing in the axiom \cite{KalyanpurPSG06}. An entity can be a class name, a property name or an individual name.  In this paper, we compute the reference counts for an axiom based on all axioms in a reliable ontology. Since it is often desired to have a reliable ontology whose axioms are more relevant to some extent, this strategy selects those axioms with the lowest score from each R-MIPS.
\end{itemize}


\begin{table}[t]
	\begin{center}
		\begin{footnotesize}
			\begin{tabular}{|l|r|l|r|l|r|}
				\hline
				Single&Time&Single&Time&Single&Time\\
				Ontolgies&(ms)&Ontolgies&(ms)&Ontolgies& (ms)\\
				\hline\hline
				km0&6373&M0&270&O0&3813 \\
				km1&6343&M1&450&O1&3089 \\
				km2&6396&M2&390&O2&7432\\
				km3&6473&M3&331&O3&3918 \\
				km4&6536&M4&322&O4&6286\\
				km5&6476&M5&329&O5&2344\\
				km6&6412&M6&502&O6&3109\\
				km7&6605&M7&477& n.a. & n.a.\\
				km8&6600&M8&556& n.a.& n.a.\\
				km9&6387&M9&328&  n.a.& n.a.\\		
				\hline
			\end{tabular}
			\vspace{3mm}
			\caption{Time in milliseconds to compute vectors for the sentences translated from the axioms in a single ontology, where `n.a.' indicates ``not applicable". \label{tab:time-pre}}
		\end{footnotesize}
	\end{center}
\end{table}

\subsection{Experimental Results}

In this section, we first present the results about the preparation about computing vectors, R-MUPS and R-MIPS. We then describe the results about revision based on all R-MIPS and local R-MIPS. Finally, a brief discussion of all experimental results is provided. 

\subsubsection{Results about preparation}\label{sec:res-pre}
Before revising a rebuttal ontology by applying one of our semantics-based scoring function, the vectors of all axioms in an ontology needs to be calculated first. This process can be done offline. Table \ref{tab:time-pre} provides the time to compute vectors for a single ontology. 

From the table, we can obviously observe that the efficiency of computing vectors mainly relies on the number of axioms. Each \textsf{km} ontology contains 1000 axioms, and thus they spent similar time (i.e., about 6 seconds). For ontologies O2 and O4, they also contain around 1000 axioms and took similar time to finish the computation of vectors. As for the ontologies obtained by translating ontology alignments, they usually include no more than 50 axioms and each of them took no more than 0.5 seconds.

\subsubsection{Results about computing all R-MIPS}
In this section, we present the results about all R-MIPS  in a rebuttal ontology with respect to its corresponding reliable ontology together with R-MUPS information, since all R-MIPS are obtained based on all R-MUPS of all unsatisfiable concepts. 

Table \ref{tab:time_computeMIPS} gives the details about the results of R-MUPS and R-MIPS. The 2nd column displays the average number of R-MUPS per unsatisfiable concept. The 3rd and 4th columns present the maximal and minimal number of R-MUPS separately. 
The columns from 6 to 8 provide average, maximal and minimal sizes of a R-MIPS. The last column describes the consumption time to compute all R-MIPS for an ontology, which includes the time to compute all R-MUPS for the considered concepts.
In the last column, the time in bold means the computation of all R-MIPS cannot be finished successfully within limited memory. In such a case, we execute the code again to find all R-MUPS for the remaining unsatisfiable concepts. In this way, all R-MIPS could be calculated based on all R-MUPS of all unsatisfiable concepts, and the revision algorithm based on all R-MIPS could be applied. This process may be performed  many times until all R-MUPS for all unsatisfiable concepts in a rebuttal ontology were found. Specifically, the code was performed twice for \textsf{KM6} (or \textsf{KM7}) and the total time is more than 300 seconds. For \textsf{OM6} and \textsf{OM8}, the process was repeated for six and seven times separately. It took more than 900 seconds to finish the computation for each of them.

\begin{table}[t]
	\begin{center}
		\begin{footnotesize}
			\begin{tabular}{|l|l|l|l|c|l|l|l|r|}
				\hline
				Pairs& \multicolumn{3}{c|}{R-MUPS Number} & R-MIPS &\multicolumn{3}{c|}{R-MIPS Size} & Time\\
				&   Avg & Max & Min & Number& Avg & Max & Min & (ms)\\
				\hline
				OM0&1.35&4&1&4&2&2&2&3711\\
				OM1&1.79&6&1&7&2&2&2&24601\\
				OM2&1.82&3&1&6&2&2&2&5926\\
				OM3&1.61&4&1&5&1.8&2&1&9230\\
				OM4&1&1&1&2&2&2&2&1190\\
				OM5&1.6&2&1&5&2&2&2&3810\\
				OM6&5.72&23&1&22&2&2&2&\textbf{969544}\\
				OM7&2&6&1&16&2&2&2&200328\\
				OM8&4.58&15&1&42&2&2&2&\textbf{974050}\\			
				OM9&2&2&2&2&2&2&2&28771\\
				\hline
				KM0&1&1&1&1&1&1&1&460\\
				KM1&1.3&2&1&1&2&2&2&18159\\
				KM2&1&1&1&4&1.25&2&1&25228\\
				KM3&1&1&1&3&1.67&3&1&52731\\
				KM4&1&1&1&2&2&3&1&38703\\
				KM5&1.12&2&1&5&1.6&3&1&36698\\
				KM6&1.02&3&1&8&1.38&2&1&\textbf{327682}\\
				KM7&1.01&2&1&8&1.88&2&1&\textbf{306044}\\
				KM8&1&1&1&3&2.67&5&1&14692\\
				
				\hline
			\end{tabular}
			\vspace{3mm}
			\caption{Time in milliseconds to compute all MIPS in an ontology, where the time in bold means it is computed by multiple runs. \label{tab:time_computeMIPS}}
		\end{footnotesize}
	\end{center}
\end{table}

From Table \ref{tab:time_computeMIPS}, we first observe that:  Although a \textsf{KM} ontology pair often contains much more unsatisfiable concepts than an \textsf{OM} pair,  each unsatisfiable concept in it has fewer R-MUPS on average, and its maximal number of R-MUPS is no more than 3. While for 60\% \textsf{OM} pairs, the maximal number of R-MUPS is larger than 3. For instance, an unsatisfiable concept in \textsf{OM6} has 23 R-MUPS at most, and 15 for  \textsf{OM8}.
That's one main reason causing that an \textsf{OM} pair spent much more time than a \textsf{KM} pair to compute all R-MIPS. Take pairs \textsf{OM7} and \textsf{KM5} as examples. It took \textsf{OM7} about 200 seconds while no more than 40 seconds for \textsf{KM5}.
Except for the maximal or average number of R-MUPS, the expressivity, total size of a pair and number of unsatisfiable concepts are also  main reasons to influence the efficiency of computing R-MIPS. For example, it took more time for the pairs \textsf{KM2}-\textsf{KM5}  than the pairs \textsf{OM2}-\textsf{OM5}, because they contain more axioms and unsatisfiable concepts. 

\begin{figure*}[!p]
	\centerline{\includegraphics[width=0.99\linewidth]{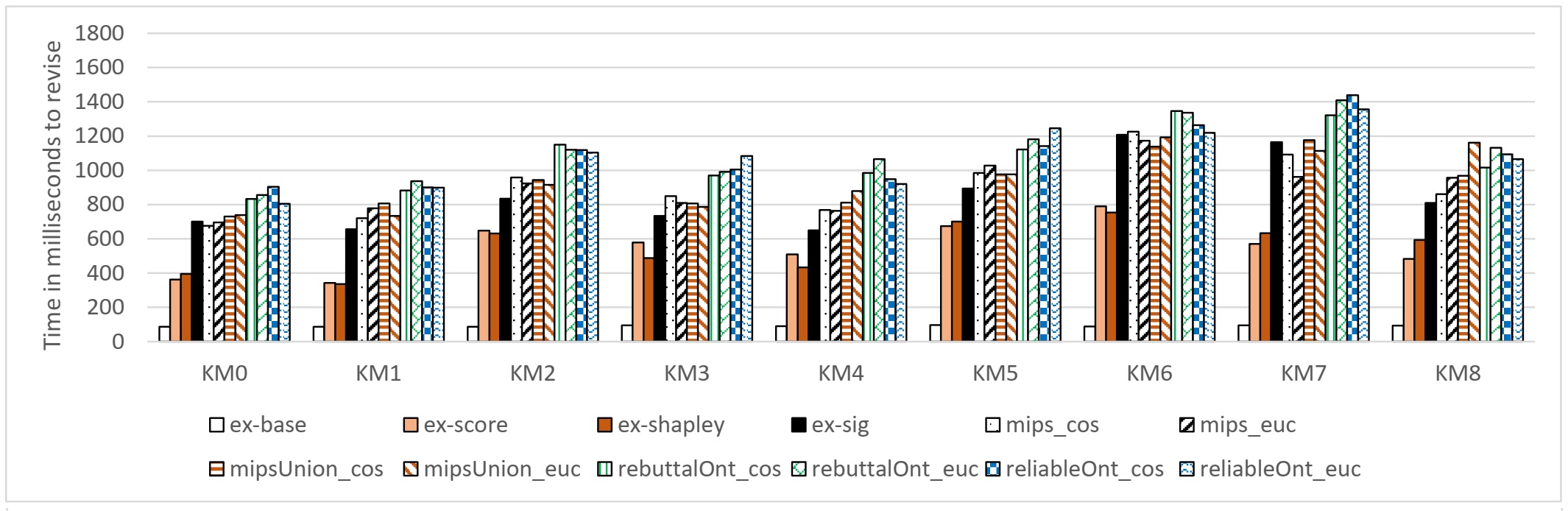}}
	\centerline{\includegraphics[width=0.99\linewidth]{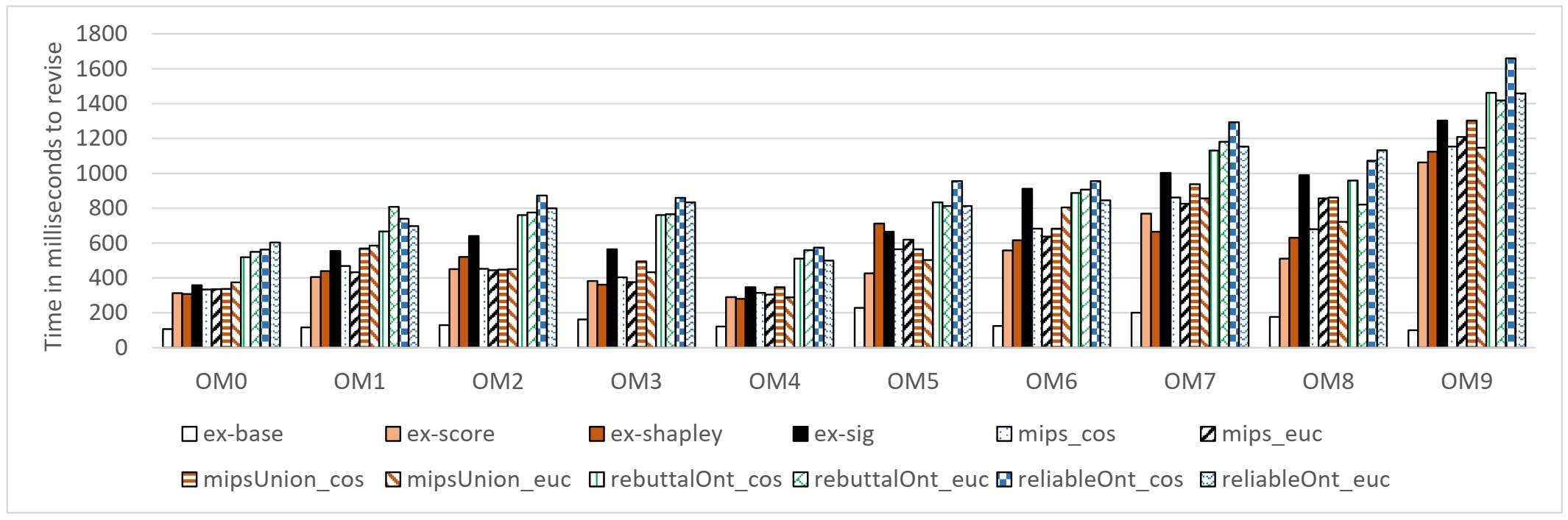}}
	\centerline{\includegraphics[width=0.99\linewidth]{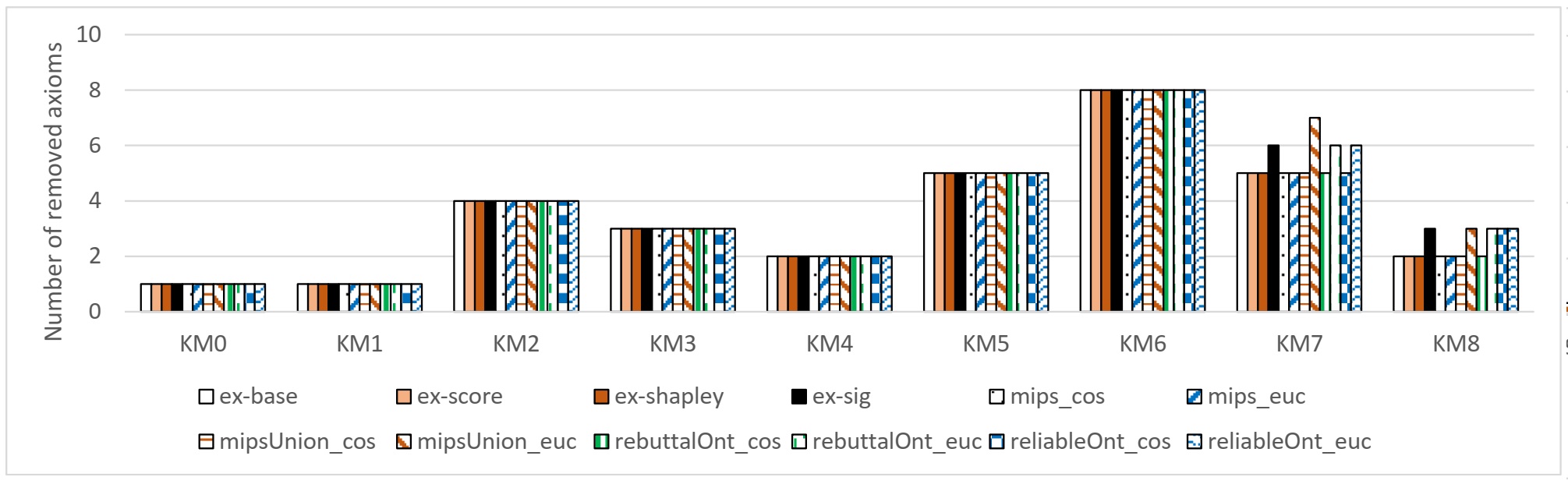}}
	\centerline{\includegraphics[width=0.99\linewidth]{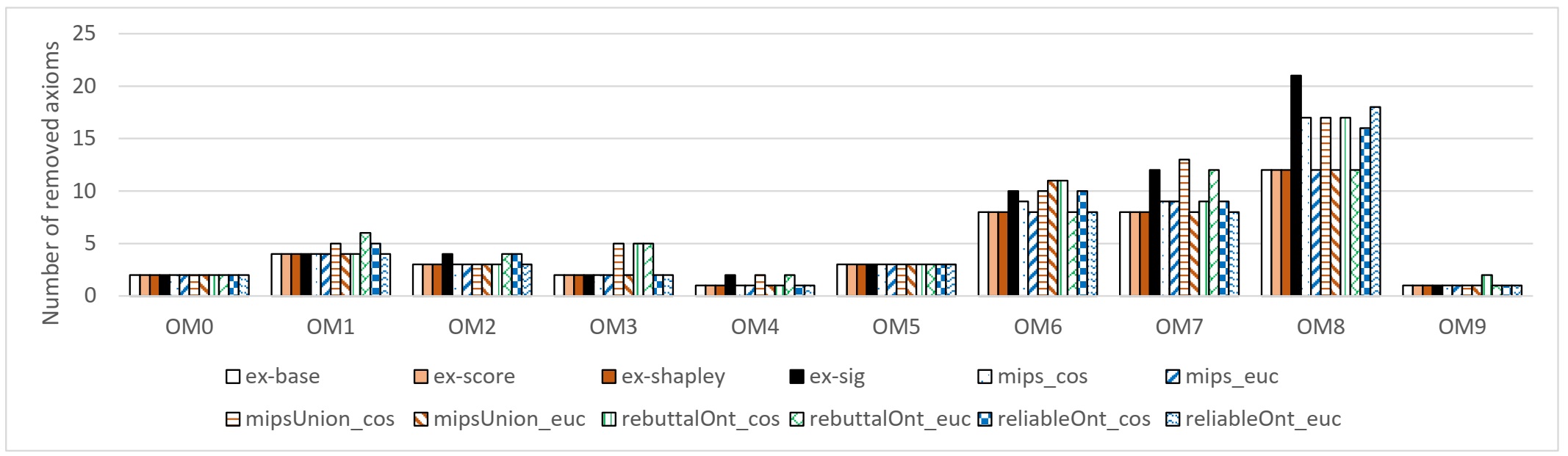}}
	\caption{Experimental results to revise an ontology pair (see X-axis) based on all R-MIPS.}\label{fig:revise-allmips}
\end{figure*}

\begin{figure*}[t]
	\centerline{\includegraphics[width=0.99\linewidth]{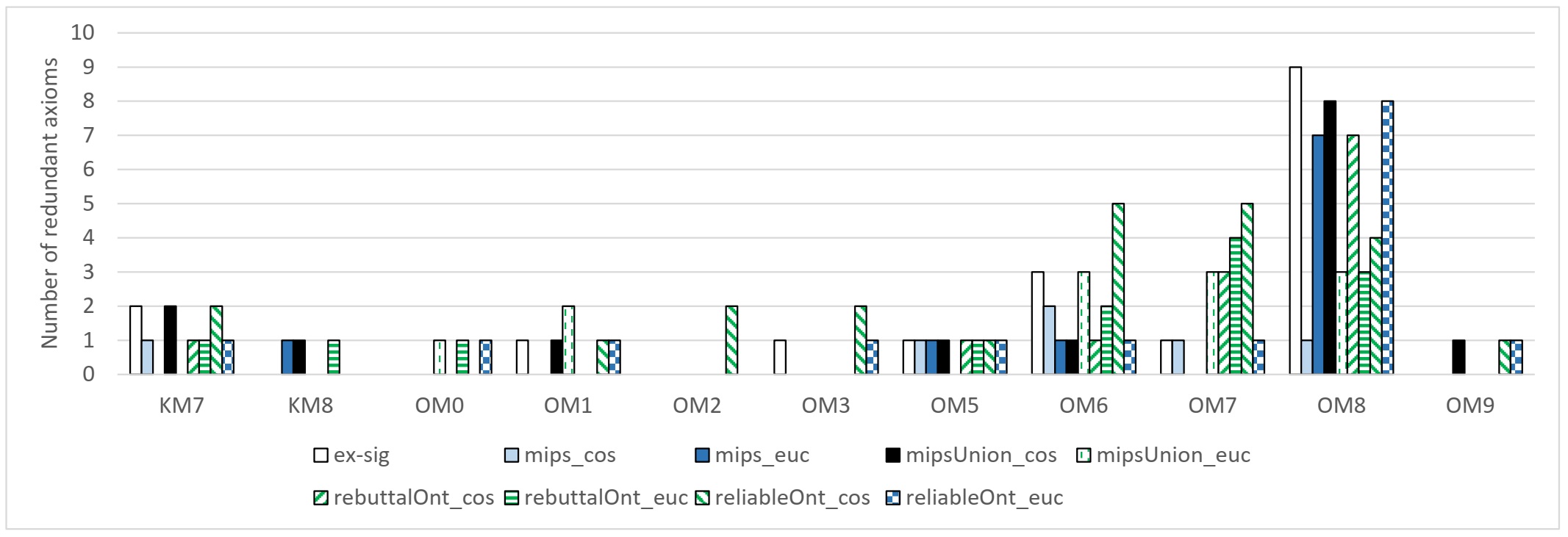}}
	\caption{Number of redundant axioms found by a revision algorithm based on all R-MIPS. }\label{fig:revise-allmips-redundant}
\end{figure*}

\subsubsection{Results about ontology revision based on all R-MIPS}
In this section, we evaluate our revision algorithms based on R-MIPS by comparing with existing ranking strategies or existing algorithms. 
Specifically, eight revision algorithms based on pre-trained models are obtained from Algorithm \ref{alg:reviseAlg} by applying different scoring functions defined in this paper and similarity measures. Another four revision algorithms are obtained by replacing the extraction strategy in Algorithm \ref{alg:reviseAlg} with \textsf{ex-base}, \textsf{score}, \textsf{ex-shapley} and \textsf{ex-sig}.
For simplicity, each revision algorithm is named by its extraction strategy in this section.

It should be noted that, when applying the extraction strategy \textsf{ex-base} to Algorithm \ref{alg:reviseAlg} proposed in this paper, we ignore subset extraction and redundancy checking as a diagnosis is computed based on all R-MIPS directly and no redundant axioms exist. This algorithm is actually the same as the revision algorithm given in \cite{Ji19access}.  In addition, when applying \textsf{ex-score} to Algorithm \ref{alg:reviseAlg}, the obtained algorithm can be seen as an enhanced version of the revision algorithm given in \cite{qi2008kernel}. The main difference is that the original revision algorithm in \cite{qi2008kernel} applies a hitting set tree algorithm \cite{reiter87} to compute a diagnosis while ours uses ILP due to its high efficiency \cite{Ji19access,JiSeu20}.

All of these algorithms were evaluated with respect to the efficiency and number of removed axioms (see Figure \ref{fig:revise-allmips}). 
The time presented in Figure \ref{fig:revise-allmips} does not include the time to calculate all R-MIPS since these algorithms are all based on all R-MIPS and we only compare their difference.

From Figure \ref{fig:revise-allmips} we can see the algorithms are all efficient to compute a diagnosis based on all R-MIPS, and usually spent no more than 14 seconds. Especially, \textsf{ex-base} is the most efficient one and often spent less than 0.2 seconds since it is not necessary to check redundancy of axioms. In fact, except \textsf{ex-base}, all algorithms spent most of their time checking redundancy of axioms.
Among the four existing algorithms, \textsf{ex-sig} is more time-consuming than others. It needs to compute reference counts for an axiom to be ranked based on all axioms in a reliable ontology, while others rank an axiom only based on all R-MIPS which usually involve fewer axioms. 
Among our eight algorithms, \textsf{mips\_cos}, \textsf{mips\_euc}, \textsf{mipsUnion\_cos} and \textsf{mipsUnion\_euc} often outperform others since fewer axioms need to be considered when ranking axioms. Comparing two similarity measures ${sim}_{cos}$ and ${sim}_{euc}$, the algorithms with the same extraction strategy but different similarity measures behave similarly.
Comparing our algorithms with existing ones, ours usually took a little more time, especially the ones considering a rebuttal or reliable ontology, since they need to spent more time to compute similarities. Take \textsf{KM7} as an example. Existing algorithms \textsf{ex-base}, \textsf{ex-score} and \textsf{ex-shapley} took about 600 milliseconds, our algorithms took around 1100 milliseconds.

According to the number of removed axioms shown in Figure \ref{fig:revise-allmips}, it can be observed that different revision algorithms removed similar number of axioms, especially for \textsf{km} ontologies. 
It should be noted that \textsf{ex-sig} removed slightly more axioms than other existing algorithms.
For example, \textsf{ex-sig} removed 10, 12 and 21 axioms for ontology pairs \textsf{OM6},  \textsf{OM7} and \textsf{OM8}, while other existing algorithms removed 8, 8 and 12 axioms separately.

We also present the number of redundant axioms found by each algorithm in Figure \ref{fig:revise-allmips-redundant}. In this figure, not all ontology pairs or extraction strategies are shown as no redundant axioms were found for such a case. For example, all algorithms did not find any redundant axioms for ontology pairs from \textsf{KM0} to \textsf{KM6}, and \textsf{ex-score} and \textsf{ex-shapley} did not produce redundancy for all tested ontology pairs.
From this figure, we can observe that usually no more than 2 redundant axioms were found, and a revision algorithm is easily to produce redundancy for ontology pair \textsf{OM8} which contains much more R-MIPS and is the most challenging one to be revised. For \textsf{OM8}, \textsf{ex-sig}, \textsf{mips\_euc}, \textsf{mipsUnion\_cos}, \textsf{rebuttalOnt\_cos} and \textsf{reliableOnt\_euc} found more than 6 redundant axioms.

\subsubsection{Ontology revision results based on local R-MIPS}\label{sec:res-localmips}

In this section, we discuss the experimental results of the adapted revision algorithms obtained by using various extraction strategies. Namely, an adapted algorithm is obtained by replacing the subset extraction strategy in Algorithm \ref{alg:adaptedReviseAlg} with one the strategies mentioned in Section \ref{sec:setting}. Similar to the revision algorithms based on all R-MIPS, an adapted algorithm is also named by the name of its extraction strategy for simplicity. 
%

\begin{figure*}[!p]	
	\includegraphics[width=0.95\linewidth]{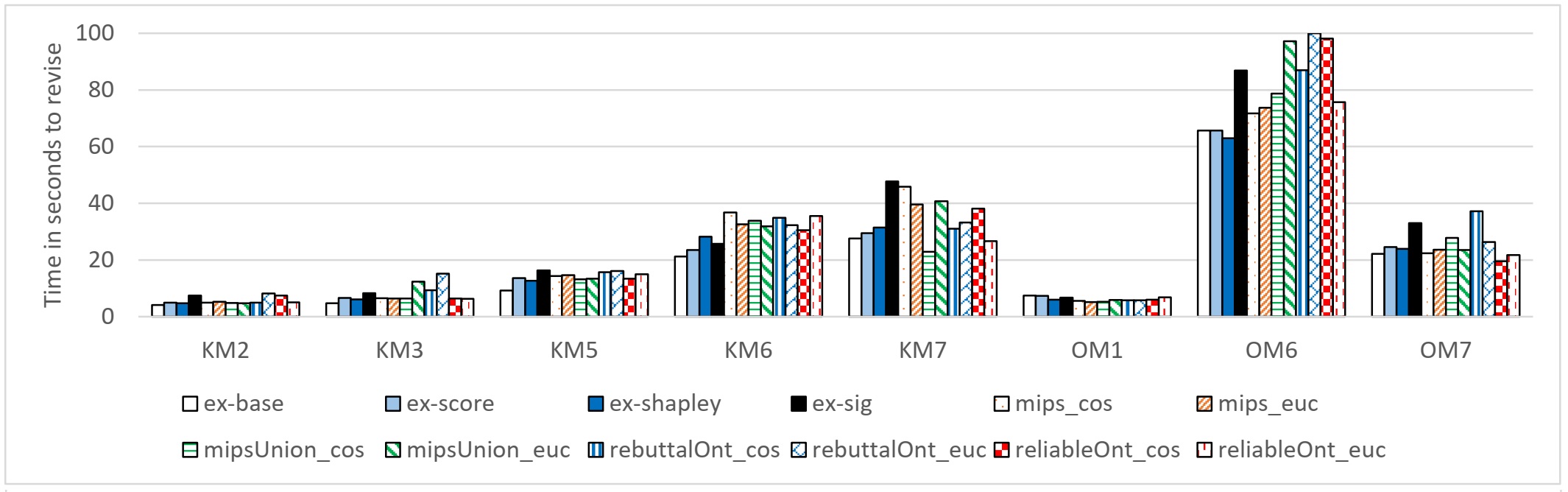}
	\includegraphics[width=0.95\linewidth]{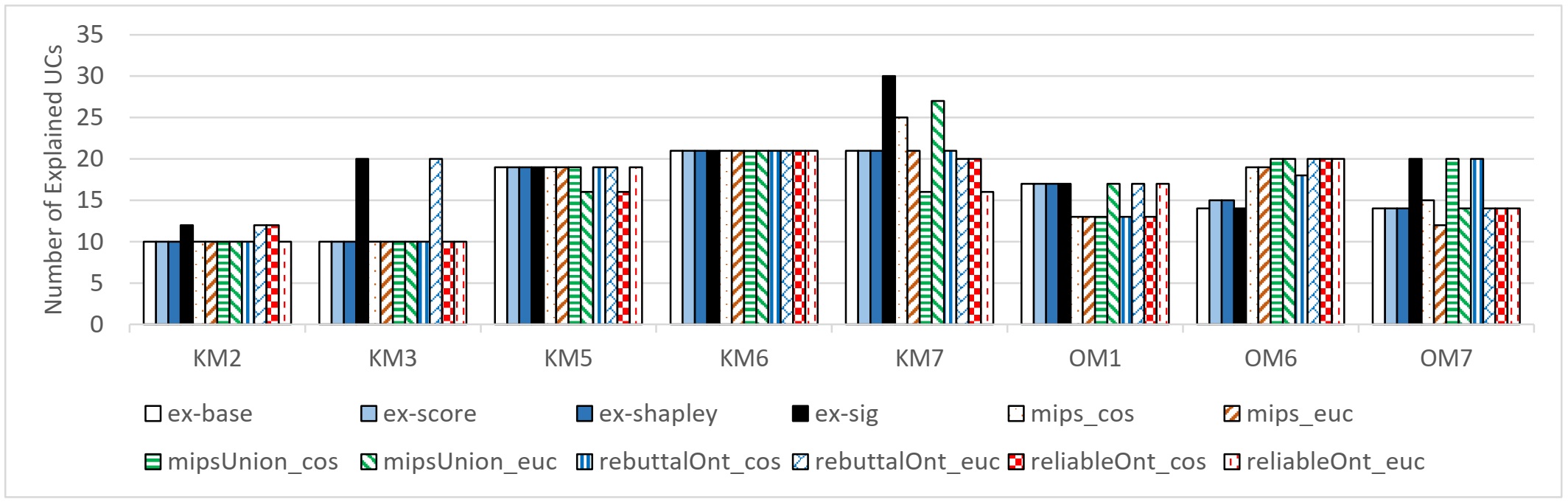}
	\includegraphics[width=0.95\linewidth]{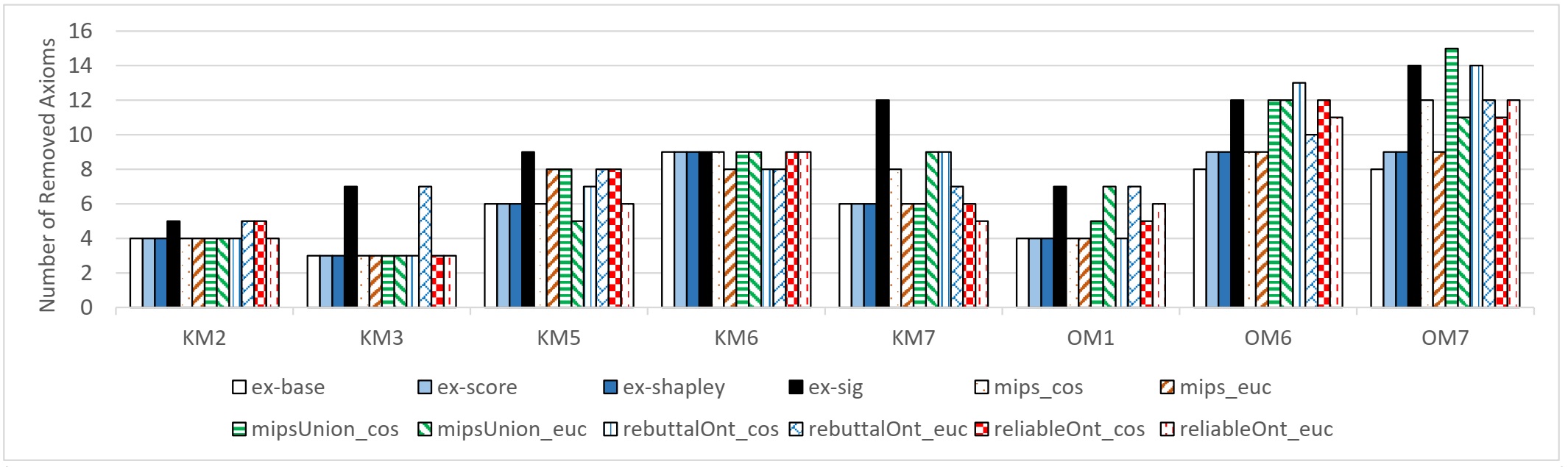}
	\caption{Experimental results of adapted revision algorithms with  step length of $10$.}\label{fig:res-revise-localMips}	
\end{figure*}

\begin{figure*}[!p]
	\includegraphics[width=0.98\linewidth]{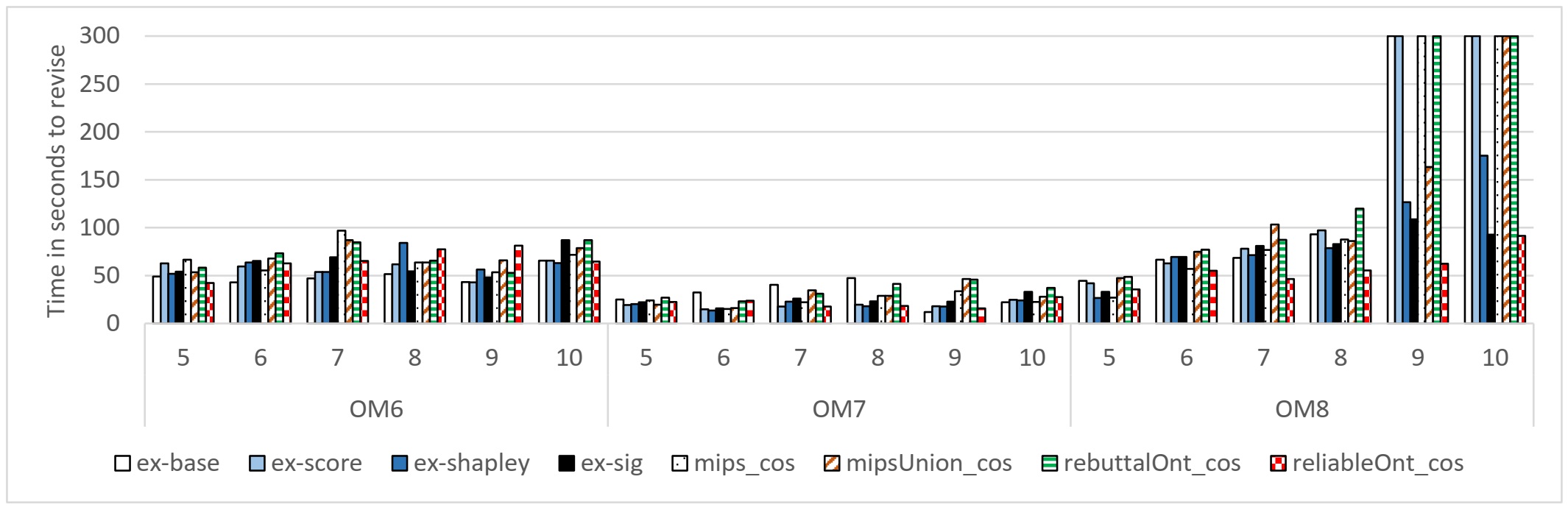}
	\includegraphics[width=0.98\linewidth]{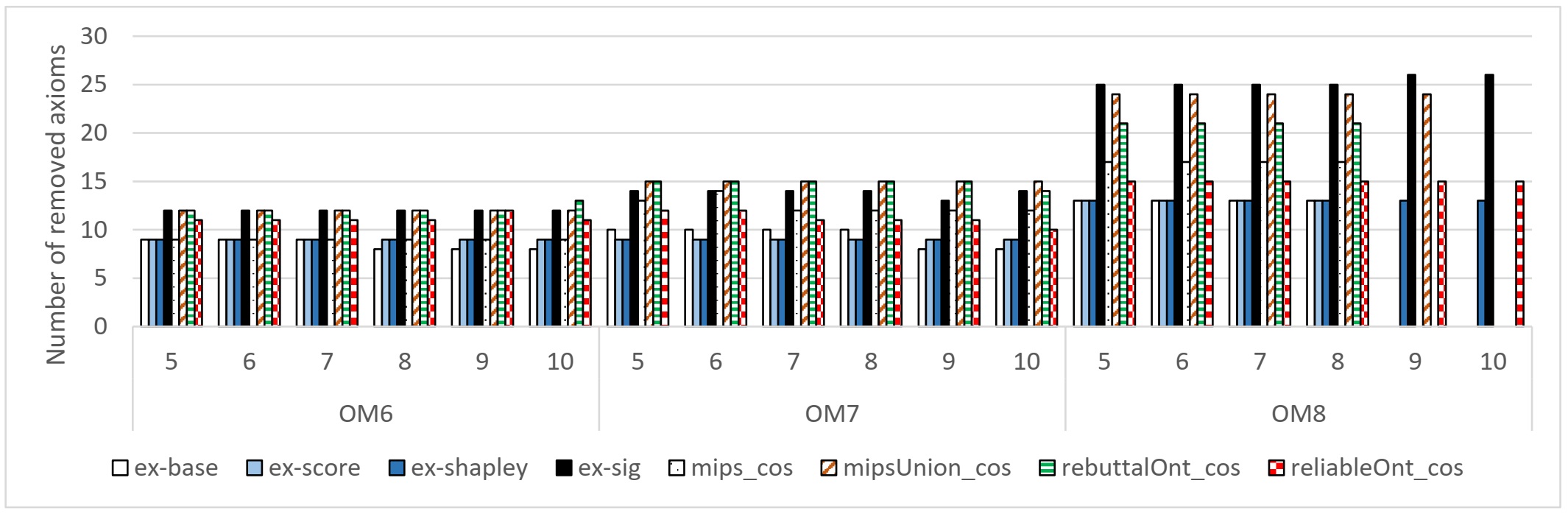}	
	\includegraphics[width=0.98\linewidth]{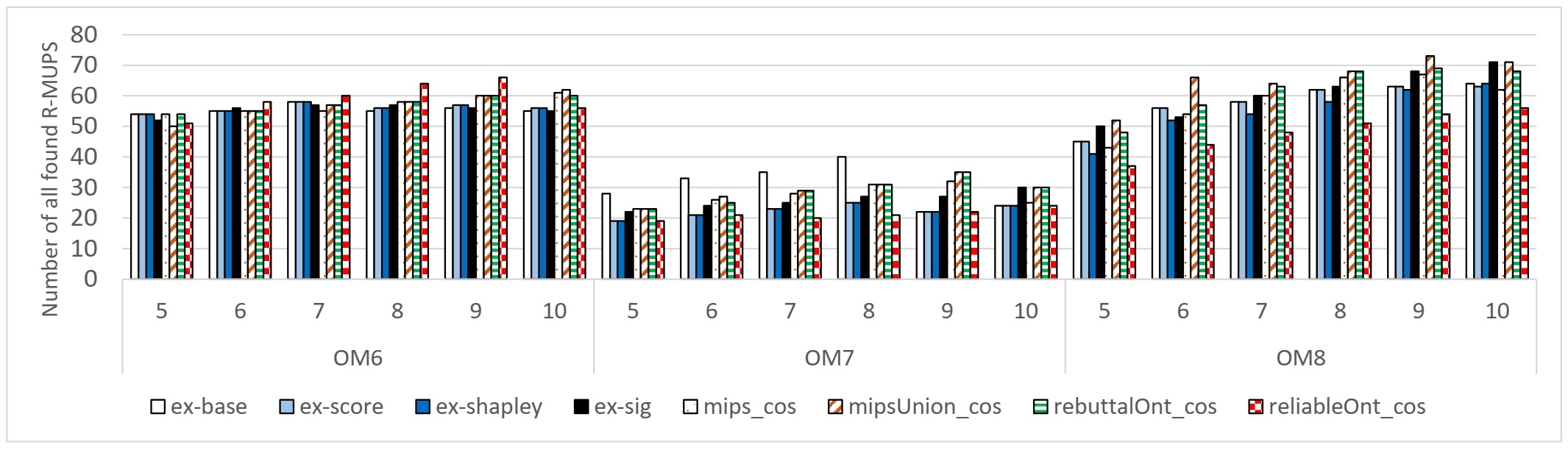}	
	\caption{Experimental results of adapted revision algorithms with various step lengths.}\label{fig:res-revise-groups}
\end{figure*}

We first conducted an experiment by setting the step length to $10$. We selected those ontology pairs with more than 30 unsatisfiable concepts since computing all R-MIPS for such a pair may be more challenging. In addition, although ontology pair \textsf{OM8} contains 84 unsatisfiable concepts, we did not test it in this experiment because all adapted algorithms failed to compute a diagnosis for it within limited memory. In this way, eight ontology pairs were selected (see Figure \ref{fig:res-revise-localMips}).
The time presented in this figure includes both the time to compute local R-MIPS and the time to compute a diagnosis.

From Figure \ref{fig:res-revise-localMips}, we obtain the following main observations: 
\begin{itemize}
	
	\item  \textsf{ex-sig} still cannot perform well within the adapted revision framework. Take ontology pair \textsf{KM7} as an example. \textsf{ex-sig} took about 50 seconds and removed 12 axioms in total, while nearly all other algorithms removed no more than 9 axioms within 40 seconds.
	
	\item The adapted algorithm with baseline extraction strategy \textsf{ex-base} may not always remove cardinality-minimal axioms. For instance, when revising ontology pair \textsf{KM5}, \textsf{ex-base} removed 6 axioms while our algorithm \textsf{mupsUnion\_euc} removed 5 axioms. It is mainly caused by the fact that each final diagnosis consists of multiple local diagnoses, and different adapted algorithms may have distinct local diagnoses. Although each final diagnosis is minimal, the baseline adapted algorithm \textsf{ex-base} may not find a cardinality-minimal diagnosis.

	\item  Comparing the number of explained unsatisfiable concepts in this figure with the total number of unsatisfiable concepts given in Table \ref{tab:ontPairs}, the adapted revision algorithms only need to explain much fewer unsatisfiable concepts than the algorithms based on all R-MIPS. For all of these selected ontology pairs, no more than 30 unsatisfiable concepts for each pair need to be explained.
	
	\item  Comparing the number of removed axioms in this figure and that in Figure \ref{fig:revise-allmips}, there is no obvious difference. For instance, all revision algorithms removed 8 axioms for \textsf{KM6}, and no more than 9 axioms were removed by the adapted ones. It shows that the adapted algorithms do not cause too much information loss.

	\item  Since computing R-MUPS and R-MIPS is the most resource-consuming step in a revision process, we compare the revision time in Figure \ref{fig:res-revise-localMips} with the explanation time of computing all R-MIPS in Table \ref{tab:time_computeMIPS}. It can be obviously seen that the adapted algorithms are much more efficient. For example, it took nearly 1000 seconds to revise \textsf{OM6} based on all R-MIPS while no more than 100 seconds for the adapted algorithms. Namely, about 90\% time was saved.
	
\end{itemize}
It needs to be mentioned that the explanation time may be influenced by many factors such as number of unsatisfiable concepts,  number and size of R-MUPS per unsatisfiable concept and the expressivity (see the analysis in \cite{Ji14kbs}).

Furthermore, to see the performance of the adapted algorithms with different step lengths, we chose the most challenging ontology pairs \textsf{OM6}, \textsf{OM7} and \textsf{OM8} to test, and varied the step length from 5 to 10. The consumption time and number of removed axioms are presented in Figure \ref{fig:res-revise-groups}. In this figure, we do not show the number of removed axioms for those ontology pairs that an algorithm failed to finish their revision processes within limited memory, and set their consumption time to be 300 seconds. Additionally, we did not test the algorithms with {sim}$_{euc}$ as a similarity measure since there is no big difference between {sim}$_{euc}$ and {sim}$_{cos}$.
From Figure \ref{fig:res-revise-groups}, we obtain the following observations:
\begin{itemize}
	\item For each ontology pair, the revision algorithms with the same configuration but different step lengths removed similar number of axioms. For example, when the steps range from 5 to 10, \textsf{ex-sig} removed 25 or 26 axioms for \textsf{OM8}, and \textsf{mips$_{cos}$} removed 15 axioms. This reflects that the changing step length leads to a slight effect on the number of axioms removed.
	
	\item The greater the step length, the more difficult it is to revise an ontology pair. Take the most challenging ontology pair \textsf{OM8} as an example. All revision algorithms successfully revised this pair when the step length $n$ is less than 9. When $n=9$, \textsf{ex-base}, \textsf{ex-score}, \textsf{mips$_{cos}$} and \textsf{rebuttalOnt$_{cos}$} failed. When $n=10$, only \textsf{ex-shapley}, \textsf{ex-sig} and \textsf{reliableOnt$_{cos}$} can revise the ontology pair successfully.  
	
	\item  The consumption time increases with the increase of the step length for the adapted algorithms with the same configuration but different step lengths. This is mainly caused by the computation of R-MUPS. Since computing R-MUPS is the most time-consuming step during a revision process, the consumption time varies when the number of all found R-MUPS varies (see the third figure in Figure \ref{fig:res-revise-groups}). 
	
	\item Comparing our four algorithms, \textsf{reliableOnt\_cos} is the most efficient one and removes the least number of axioms. Take the ontology pair \textsf{OM8} as an example. When the step length is 8, \textsf{reliableOnt\_cos} spent 55 seconds and removed 15 axioms, but other three algorithms spent more than 85 seconds and removed more than 17 axioms.  This reflects that removing different axioms may affect the efficiency of a revision algorithm to a great extent.
	
\end{itemize}

\subsubsection{Discussion of Experimental Results}
Based on the analysis given in the sections from \ref{sec:res-pre} to \ref{sec:res-localmips}, we provide a brief discussion here to help readers grasp the main conclusions of our experiments and summary guidelines to choose different revision algorithms.

Firstly, the adapted algorithms are much more efficient than those considering all R-MIPS. Furthermore, although the adapted algorithms are based on local R-MIPS and may remove more axioms, the difference is minor. Of course, the adapted algorithms may fail to revise a rebuttal ontology with too many R-MUPS for some unsatisfiable concepts either. In such cases, it is not suitable to revise ontologies based on all R-MIPS or local R-MIPS. This problem will be studied in the future.

Secondly, our adapted algorithms with different step lengths present promising results, especially the algorithm based on a reliable ontology \textsf{reliableOnt\_cos}. For the adapted  algorithms, 
a suitable step length is critical, and it may be varied when the tested ontologies change. According to our observations, a step length could be set according to the expressivity, number of unsatisfiable concepts and size of an ontology pair. For example, if the considered ontologies are expressive or many unsatisfiable concepts are involved, a lower length is preferred.

Thirdly, for the adapted revision algorithms with the same configuration but different extraction strategies, they may remove similar number of axioms while the removed axioms are different. This may cause that distinct unsatisfiable concepts need to be explained, and the efficiency of an revision algorithm will be influenced accordingly.

Finally, the users could select an extraction strategy according to  its efficiency, number of removed axioms, the characteristics of ontologies, or semantics. For example, if a user prefers to choose an efficient algorithm considering semantics, or keep those rebuttal axioms that are more relevant to the reliable ones, \textsf{reliableOnt\_cos} or \textsf{reliableOnt\_euc} could be selected. If those axioms that are less relevant to the axioms in R-MIPS are preferred to be kept, \textsf{mips\_cos}, \textsf{mips\_euc}, \textsf{mipsUnion\_cos} or \textsf{mipsUnion\_cos} are good choices. If removing less axioms is the most important thing, \textsf{ex-base} is recommended. 

\section{Related Works}\label{sec:related}

In this section, we discuss the related works on ontology revision and ontology mapping revision.

\subsection{Ontology revision approaches}
Ontology revision approaches can be generally divided into automatic and interactive approaches. For those theoretical ones like \cite{ribeiro2021revising}, revision works without considering logical conflicts such as \cite{PesquitaC12,cardoso2018supporting}, and axiom-weakening approaches \cite{MicalizioP18}, we recommend readers to read the relevant references.

Automatic ontology revision approaches resolve incoherence or inconsistency without people's participation.
The authors in \cite{qi2008kernel} proposed a kernel revision operator to deal with incoherence, and assigned scores to axioms by considering their weights or frequencies in all R-MIPS.
Similarly, the authors in \cite{golbeck2009trust} also defined a kernel revision operator and incision functions, but they dealt with inconsistency and exploited trust information to choose axioms.
To improve the efficiency of revising ontologies, the authors in \cite{fu2016graph} focused on DL-Lite, and converted DL-Lite ontologies into graphs. They computed the MIPS from minimal incoherence-preserving path-pairs, and scored an axiom according to its logical closure. 
The work in \cite{JiSeu20} also followed the idea of defining a kernel revision operator, but its novelty lies in considering a partial order of axioms to stratify axioms and selecting axioms based on integer linear programming.
In addition, there are some works to revise ontologies by defining new semantics. For instance, the authors in \cite{wang2015instance,ZhuangWWQ16} focused on DL-Lite ontologies and defined type semantics instead of standard DL semantics.
In this paper, we also exploit a kernel revision operator like existing works. One main difference is that we consider semantic similarity between axioms to define incision functions based on a pre-trained model. The other main difference is that we design various revision algorithms considering the semantic similarity, especially the adapted ones which are trade-off to balance the revision of all unsatisfiable concepts at one time or one by one.

An interactive ontology revision approach needs the participation of users to make some decisions. According to our investigation, most of the existing ontology revision approaches focus on resolving incoherence or inconsistency automatically, and few of them proposed interactive algorithms. One typical interactive ontology revision approach was introduced in \cite{nikitina2012interactive}.
This work separated a DL knowledge base into two parts: one including the axioms that should be inferred (marked as $M_1$) and the other containing those that cannot be inferred  (marked as $M_2$). Its goal is to find a complete and consistent revision state which makes any axiom in $M_2$ cannot be inferred by $M_1$. The revision process is interactive and displays one unlabeled axiom each time to a user for deciding whether to accept it or not.
This process is repeated until all unlabeled axioms are labeled.
The scoring function in this work was defined as the number of axioms in an ontology that can be inferred by a specific axiom set.
It relies on logical reasoning which is always resource-consuming. Our scoring functions are based on pre-trained models and independent of any logical reasoner. The most resource-consuming part of them is to calculate vectors by using a pre-trained model which can be done offline.

\subsection{Ontology mapping revision approaches}
Ontology revision is closely related to ontology mapping revision. Most approaches of ontology revision can be applied to revise ontology mappings. However, ontology mappings have their own characteristics so that various approaches to revising ontology mappings have been designed. Such approaches can also be divided into automatic and interactive ones. We mainly focus on those works where a scoring function is defined.

Among the works about interactive ontology mapping revision, the authors in \cite{Meilicke08mapping} employed a reasoning-based approach to locating the incoherence of mappings and defined a bridge rule function to
rank the impacts of mappings, which can reduce the number of decisions made by an expert.
The work in \cite{li2023graph} followed the approach proposed in the work of interactive ontology revision in \cite{nikitina2012interactive}. One main difference is that the work in \cite{li2023graph} focused on DL-Lite ontologies and transferred them into a graph for improving efficiency. Another main difference is that the scoring function defined in \cite{li2023graph} considered both the number of mapping arcs in a specific set and the weights of mappings.

As for the approaches of automatic ontology mapping revision, an early work was proposed in \cite{qi2009conflict} which defined a conflict-based revision operator and designed two specific algorithms to instantiate this operator. One algorithm stratified the axioms in an alignment according to their weights, and the other utilized a signature-relevance selection function to distinguish axioms.
The ontology matching system LogMap \cite{logmap} exploited horn propositional logic to model unsatisfiable concepts and the incoherence of mappings, and repaired mappings by removing one axiom with the lowest weight from each MIPS. The matching system AMLR \cite{amlr} also removed axioms according to their weights and other heuristics.
Some other ontology matching systems like ELog \cite{elog} and PDLMV \cite{pdlmv} employed probabilistic reasoning techniques based on the weights of mappings. 

Obviously, the ranking strategies used in these existing works mainly depend on weights, logical reasoning, trust information or rankings obtained according to ontology syntax. Few of them consider the semantics of axioms. This problem has been addressed in our previous work \cite{JiAs23}, but it only considers repairing a single ontology. This work deals with the task of ontology revision so that different scoring functions and algorithms were proposed.

\section{Conclusion and Future Works}\label{sec:conclusion}
In this paper, we first defined four scoring functions to rank an axiom in R-MIPS by considering its semantic relationship with axioms in R-MIPS, all rebuttal axioms or all reliable axioms. The semantic relationship between two axioms is measured by the similarity between their corresponding vectors.
We then proposed a pre-trained model-based ontology revision algorithm by considering all R-MIPS, and then an adapted algorithm was designed to deal with those challenging ontology pairs that it is hard to compute all of their R-MIPS within limited resources. The adapted algorithm relies on local R-MIPS computed based on all R-MUPS of some unsatisfiable concepts. 
We implemented our algorithms and evaluated them with 19 ontology pairs coming from real-life ontologies. The experimental results reflect that the adapted algorithms are very efficient and could save at most about 90\% of the time for some tested ontology pairs. Among our algorithms based on pre-trained models, \textsf{relibleOnt\_cos} outperforms others in many cases.
We also provided a brief discussion about the overall experimental results to conclude main observations, and provided several guidelines for users to choose different algorithms.

In the future, we first plan to study how to deal with the cases that an unsatisfiable concept in a rebuttal ontology contains too many R-MUPS. In such a case, it is hard to compute all R-MUPS within limited time and memory, and both algorithms presented in this paper may not be applied. Instead of computing all R-MUPS, we will compute some R-MUPS using a depth-first search strategy or a predefined number of R-MUPS. 
Secondly, we will evaluate the reasoning ability of some Large Language Models like ChatGPT which is an advanced AI language model developed by OpenAI for natural language processing and conversation \cite{chatgpt}. Its key advantage lies in its proficiency in understanding and generating human-like text, making it a valuable foundation for a wide range of natural language processing tasks and applications. Thus, we will explore the reasoning capability of ChatGPT and employ it to select axioms for removing.

\section*{Acknowledgements}
This work was partially supported by the CACMS Innovation Fund(CI2021A00512), the Fundamental Research Funds for the Central Universities, JLU, the Natural Science Foundation of China grants (U21A20488, U19A2061, 42050103 and 62076108) and the Fundamental Research Funds for the Central Public Welfare Research Institutes undergrant(ZZ140319-W).


\bibliographystyle{plain}


\end{document}